\documentclass[final]{article}
\PassOptionsToPackage{round}{natbib}

\usepackage{neurips_2024}

\usepackage[utf8]{inputenc} 
\usepackage[T1]{fontenc}    
\usepackage{hyperref}       
\usepackage{url}            
\usepackage{booktabs}       
\usepackage{amsfonts}       
\usepackage{nicefrac}       
\usepackage{microtype}      
\usepackage{xcolor}         
\usepackage{graphicx}
\usepackage{amsmath} 
\usepackage{multirow}
\usepackage{subcaption}
\usepackage{algorithm}
\usepackage{algorithmic}
\usepackage{natbib}

\title{Balance of Number of Embedding and their Dimensions in
Vector Quantization}

\author{Hang Chen\\
  BeiJing JiaoTong University\\
  \texttt{22120028@bjtu.edu.cn} \\
  \And
  Sankepally Sainath Reddy \\
  Indian Institute of Technology \\
  \texttt{sankepallysainathreddy@gmail.com} \\
  \AND
  Ziwei Chen\\
  BeiJing JiaoTong University \\
  \texttt{zwchen@bjtu.edu.cn } \\
  \And
  Dianbo Liu \\
  National University of Singapore \\
  \texttt{dianbo@nus.edu.sg} \\
}

\begin{document}

\maketitle

\begin{abstract}

The dimensionality of the embedding and the number of available embeddings ( also called  codebook size) are critical factors influencing the performance of Vector Quantization(VQ), a discretization process used in many models such as the Vector Quantized Variational Autoencoder (VQ-VAE) architecture. This study examines the balance between the codebook sizes and dimensions of embeddings in VQ, while maintaining their product constant. Traditionally, these hyper parameters are static during training; however, our findings indicate that augmenting the codebook size while simultaneously reducing the embedding dimension can significantly boost the effectiveness of the VQ-VAE. As a result, the strategic selection of codebook size and embedding dimensions, while preserving the capacity of the discrete codebook space, is critically important. To address this, we propose a novel  adaptive dynamic quantization approach, underpinned by the Gumbel-Softmax mechanism, which allows the model to autonomously determine the optimal codebook configuration for each data instance. This dynamic discretizer gives the VQ-VAE remarkable flexibility. Thorough empirical evaluations across multiple benchmark datasets validate the notable performance enhancements achieved by our approach, highlighting the significant potential of adaptive dynamic quantization to improve model performance.


\end{abstract}

\section{Introduction}

The Vector Quantized-Variational Autoencoder (VQ-VAE)\citep{van2017neural}, which blends the methodologies of Variational Autoencoders (VAEs) \citep{kingma2013auto} and Vector Quantization (VQ) \citep{gray1984vector}, effectively tackles the issue of noisy reconstructions commonly associated with VAEs by learning data representations that are both compact and efficient. In the VQ-VAE architecture, a vector quantization(VQ) layer converts continuous embeddings \(z_e\) into their discrete equivalents \(z_q\). Due to the non-differentiable nature of the discretization process, the straight-through estimator (STE) \citep{bengio2013estimating} is used to perform backpropagation. The codebook, which stores these discrete values, has been a focal point of significant research aimed at addressing codebook collapse, a challenge introduced by the use of STE. Innovative techniques such as the use of Exponential Moving Average (EMA)\citep{van2017neural}  and the introduction of multi-scale hierarchical structures in the VQ-VAE-2 \citep{razavi2019generating}  have proven effective in improving codebook efficiency.

A variety of techniques have been utilized to enhance the diversity of codebook selections and alleviate issues of underutilization. These methods include stochastic sampling \citep{roy2018theory, kaiser2018fast, takida2022sq}, repeated K-means \citep{macqueen1967some, lancucki2020robust}, and various replacement policies \citep{zeghidour2021soundstream, dhariwal2020jukebox}. Furthermore, studies have investigated the impact of codebook size on performance. For example, Finite Scalar Quantization (FSQ) \citep{mentzer2023finite} achieves comparable performance to VQ-VAE with a reduced quantization space, thereby preventing codebook collapse. Similarly, Language-Free Quantization (LFQ) \citep{yu2023language} minimizes the embedding dimension of the VQ-VAE codebook to zero, improving reconstruction capabilities. The VQ optimization technique \citep{li2023resizing} allows for adjustments in codebook size without necessitating retraining, addressing both computational limits and demands for reconstruction quality. Despite these significant developments, an automated approach to determine the ideal codebook size and dimension of each codeword for specific datasets and model parameters remains unexplored.


This research  tackles the fore-mentioned gap by exploring the ideal codebook size for VQ-VAE. We present three major contributions:
\begin{itemize}
\item \textbf{Optimal Codebook Size and Embedding Dimension Determination:} We systematically explored different combinations of  codebook size and embedding dimension to understand optimal balances for different tasks.
\item \textbf{Adaptive Dynamic Quantization:} We introduce novel quantization mechanism that enables data points to autonomously select the optimal codebook size and embedding dimension.
\item \textbf{Empirical Validation and Performance Enhancement:}
 We conducted thorough empirical evaluations across multiple benchmark datasets to understand behaviors of the dynamic quantizer.
\end{itemize}

\section{Preliminaries}
\label{sec2}
In this study, $x$ is the input data, $\hat{x}$ is the output data, \( E(\cdot) \) is the encoding function, \( G(\cdot) \) is the decoding function, and \( \mathcal{L}(\cdot) \) is the loss function.

\subsection{Vector-quantized networks}
\label{sec2.1}
A VQ-VAE network comprises three main components: encoder, decoder and a quantization layer \( h(\cdot) \).

\begin{equation}
\label{eq1}
\hat{x} = G(h(E(x)))= G(h(z_{e} ))=G(z_{q} )
\end{equation}
The VQ layer \(h\left ( \cdot  \right )\)  quantizes the vector \(z_{e}=E(x)\), by selecting one vector from \(N\) vectors so that $z_q=h(z_e)$. These \(N\) candidate vectors are called code embeddings or codewords, and they are combined to form a codebook \(C=\left \{ c_{1},c_{2},c_{3}...c_{N}  \right \}\). 

In codebook \(C\), each code vector has a fixed index. Therefore, for the quantization layer, the discrete embedding \(z_{q}\) that \(z_{e}\) is quantized into is selected from the \(N\) code vectors in the codebook, \(z_{q} \in R^{T\times D}\), \(D\) is the embedding dimension and \(T\) refers to the number of data points. This selection process is implemented using the nearest neighbor method:
\begin{equation}
\label{eq2}
z_{q}=c_{k} ,\,\,\,\,\,\,\,\,where\,\,k =\underset{j}{argmin}  \left \| z_{e} -c_{j}  \right \|_{2}
\end{equation}

In this process, \(P_{Z}\) is represented as the data distribution after being encoded by the encoder, and \(Q_{Z}\) represents the data distribution in the codebook space. After quantization, \(z_{q}\) is used to predict the output \(\hat{x} =G(z_{q} )\), and the loss is computed with the target \(x : L(\hat{x} , x)\). The above introduces the data flow process of a single vector passing through the VQ-VAE model. For images, VQ is performed at each spatial location on the existing tensor, where the channel dimensions are used to represent vectors. For example, the graphical data of \(z_{e}\in R^{B\times C\times H\times W}\) is flattened into \(z_{e}\in R^{(B\times H\times W)\times C}\), and then it goes through the process \( z_{e}\in R^{(B\times H\times W) \times C}\xrightarrow[projection]{h( \cdot )} z_{q}\in R^{T\times D},T=B\times H\times W \). $B$, $C$, $H$ and $W$ are the batch size, number of channels, image height, and width of image data, respectively. The standard training objective of VQ-VAE is to minimize reconstruction loss, its total loss is:
\begin{equation}
\label{eq3}
\mathcal{L} = \mathcal{L}_{task}+ \mathcal{L}_{cmt}=\left \| x-\hat{x}  \right \| _{2}^{2} +\beta \left \{  ||sg[z_{e}]-z_{q} ||_{2}^{2}+\alpha||z_{e}-sg[z_{q}]||_{2}^{2}]\right \} 
\end{equation}
Including the terms with \(sg\left [ \cdot  \right ] \) which is stop-gradient operation, in the two items are the commitment loss. The gradient through non-differentiable step is calculated using straight-through estimation \(\frac{\partial\mathcal{L}}{\partial E}=\frac{\partial\mathcal{L}}{\partial{z_{q}}}  \frac{\partial\mathbf{z}_e}{\partial E}\approx \frac{\partial\mathcal{L}}{\partial{z_{q}}}\frac{\partial{z_{q}}}{\partial{z_{e}}}  \frac{\partial\mathbf{z}_e}{\partial E}\) is accurate. 
\(\alpha \) is a hyperparameter and set to \(0.25\). \(\beta\) is a parameter that controls the proportion of quantization loss. Exponential Moving Average (EMA) is a popular approach to update the codebook according to values of the encoder outputs: \(\mathbf z_q^{(t+1)}\leftarrow(1-\gamma )\cdot\mathbf z_q^{(t)}+\gamma \cdot\mathbf  z_e^{(t)}\), where \(\gamma\) is the decay coefficient.

\subsection{Related work}\label{sec2.2}

Vector Quantization (VQ) is a prominent technique in deep learning for deriving informative discrete latent representations. As an alternative to VQ, Soft Convex Quantization\citep{gautam2023soft} solves for the optimal convex combination of codebook vectors during the forward pass process. Various methods, such as HQ-VAE\citep{takida2023hq}, CVQ-VAE\citep{zheng2023online}, HyperVQ\citep{goswami2024hypervq}, and EdVAE\citep{baykal2023edvae}, have been developed to enhance the efficiency of codeword usage, achieve effective latent representations, and improve model performance. The Product Quantizer\citep{baldassarre2023variable} addresses the issue of local significance by generating a set of discrete codes for each image block instead of a single index. One-Hot Max (OHM)\citep{lohdefink2022adaptive} reorganizes the feature space, creating a one-hot vector quantization through activation comparisons. The Residual-Quantized VAE\citep{lee2022autoregressive} can represent a \(256 \times 256\) image as an \(8 \times 8\) resolution feature map with a fixed codebook size. DVNC\citep{liu2021discrete} improves the utilization rate of codebooks through multi head discretization. The following DVQ\citep{liu2022adaptive} dynamically selects discrete compactness based on input data. QINCo\citep{huijben2024residual}, a neural network variant of residual quantization, uses a neural network to predict a specialized codebook for each vector based on the vector approximation from the previous step. Similarly, this research utilizes an attention mechanism\citep{vaswani2017attention} combined with Gumbel-Softmax to ensure the selection of the most suitable codebook for each data point.
\section{Method}
\label{sec3}
\begin{figure}
  \centering
  \includegraphics[width=1.0\textwidth]{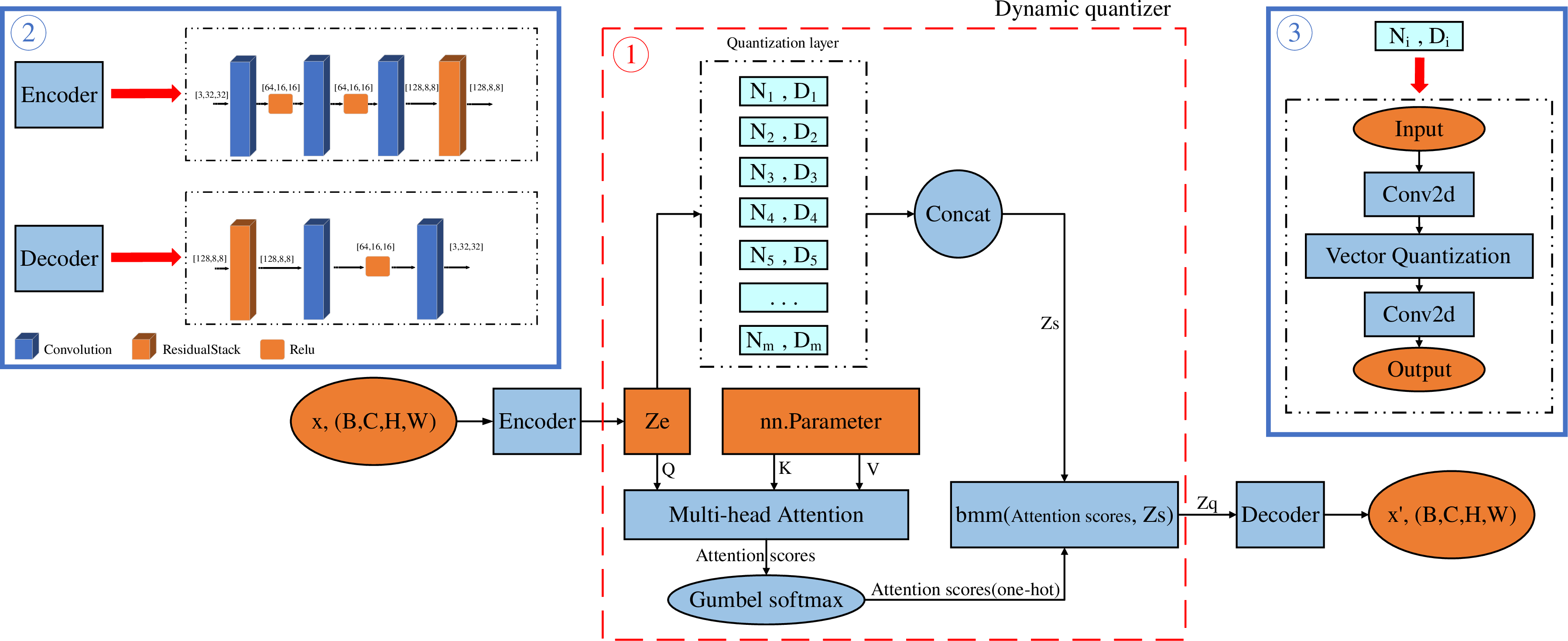}
  \caption{Schematic diagram of VQ-VAE network with adaptive dynamic quantization structure: \((1)\)Utilizes the multi-head "attention mechanism" and Gumbel-Softmax operation to achieve dynamic selection of the quantization codebook. \((2)\)Both the encoder and decoder employ CNN+ResNet architectures. \((3)\)The experimental model includes functions for dimension transformation before and after quantization within the quantization layer to facilitate data flow.}
  \label{fig1}
\end{figure}
\subsection{Codebook structure search}\label{sec3.1}

To investigate the impact of codebook size and embedding dimension on VQ-VAE's performance while maintaining a fixed capacity of the discrete information space, we conducted a series of experiments with various combinations of codebook sizes and embedding dimensions.. We denote the number of codewords by \(N\) and the embedding dimension by \(D\). For a given discrete information space capacity \(W\), the possible codebook structures are represented as \(H = \{ [N_{1}, D_{1}], [N_{2}, D_{2}], \ldots, [N_{m}, D_{m}] \}\), where \(N_{i} \times D_{i} = W\) and \(N_{i}, D_{i} \in \mathbb{N}^{+}\). \textbf{In this study, we fix $W$ and explore different combination of $N$ and $D$.}

For these codebook combinations, we ensure that the number of codewords \(N\) is always greater than the embedding dimension \(D\) for two reasons: (1) a small number of codewords can significantly degrade the model's performance. (2) maintaining consistency in subsequent experiments. These systematic experiments aim to validate our hypothesis that the codebook size influences model performance and its pattern of impact. We hypothesize that increasing the number of codewords can enhance the reconstruction performance of VQ-VAE. Based on the results obtained from exploring different codebook combinations, further refined experiments are detailed in Section \ref{sec3.2}.

\subsection{Adaptive dynamic quantization mechanism}\label{sec3.2}

Unlike methods that use a single fixed codebook, our approach utilizes multiple codebooks while ensuring that the product of the codebook size and the embedding dimension remains constant. The diagram of our experimental model is presented in Figure \ref{fig1}. The detailed algorithm of VQ-VAE with Adaptive Dynamic Quantization is clearly depicted in Algorithm \ref{algo1}.
\begin{algorithm}
\caption{VQ-VAE with Adaptive Dynamic Quantization.}\label{algo1}
\begin{algorithmic}[1]

\STATE \textbf{Input}: Original images \(x\), Number of embeddings \(num_{- } embeddings\), Embedding dimension \(embedding_{- } dims\), Hidden layer dimension \(num_{- } hiddens\)

\STATE{\textbf{Output}: Reconstructed images \(\hat{x}\), Reconstruction loss \(L\) }
\STATE {Randomly initialize \(K\)  with sizes \([m, 1, num_{- }hiddens]\). }
  
\STATE {Encoder \(E\left ( \cdot  \right )\), Decoder \(G\left ( \cdot  \right )\), Quantization layers \(h_{i} \left ( \cdot  \right ) ,i\in\left \{ 1, 2, 3, ..., m \right \} \), Gumbel-Softmax function \(GS\left ( \cdot  \right )\) }
\STATE {\(z_{e} = E(x)\) } 
\STATE {\(Q= z_{e}.flatten(-1,num_{- }hiddens) \)} 
\STATE {One-hot selection score:\(scores = GS(MultiHead(Q,K))\) } 

\STATE {\(z_{q}^{i}\),  \(vq_{- } loss_{i}   = h_{i} (z_{e})\) }
\STATE {\(z_{s}=concat\left [ z_{q}^{1},z_{q}^{2},...z_{q}^{m} \right ]\),  \(extraloss=\frac{1}{m} \sum_{i=1}^{m} vq_{- } loss_{i}\) }
\STATE {\(z_{q}=BMM(scores,z_{s})\) }
\STATE {\(\hat{x} =G\left ( z_{q}  \right )\) }
\STATE {\(recon_{- } loss=MSE(x,\hat{x} )\), \(L =recon_{- } loss+extraloss\).}

\end{algorithmic}
\end{algorithm}
A pool of \(m\) discretization codebooks \(H = \left\{ [N_{i}, D_{i}] \right\}_{i \in [m]}\) is provided for all representations undergoing discretization. Each codebook \([N_{i}, D_{i}]\) is associated with a unique quantization layer \(h_{i}(\cdot)\). These discretization codebooks and their corresponding quantization layers are independent and do not share parameters. The discrete vector obtained by quantizing \(z_e\) through the quantization layer \(h_{i}(\cdot)\) is denoted as \(z_{q}^{i}\). Each quantizer \(h_{i}(\cdot)\) possesses a unique key vector \(k_{i}\) and value vector \(v_{i}\), which are initially randomly initialized. To perform parameter feature learning for the quantization layers, we utilize the Multi-Head Attention mechanism, defined as \(\text{scores}_{\text{attention}} = \text{MultiHead}(Q, K, V)\), where \(Q\) is the flattened vector of \(z_e\) and \(K = \text{concat}(k_{1}, k_{2}, \ldots, k_{m})\), \(V = \text{concat}(v_{1}, v_{2}, \ldots, v_{m})\). The Gumbel-Softmax operation is applied to the attention scores for one-hot selection to determine which codebook to use, with the categorical distribution given by \(\pi^{z_{e}}(i) = \frac{\exp(k_{i}^{\top} z_{e})}{\sum_{j \in [N]} \exp(k_{j}^{\top} z_{e})}\).

Finally, the attention scores obtained after the Gumbel-Softmax operation are combined with multiple quantized vectors to produce the final quantized vector \(z_q\). This is achieved using batch matrix multiplication, denoted as \(z_q = \text{BMM}(\text{scores}_{\text{attention}}, z_s)\), where \(z_s = \text{concat}(z_{q}^{1}, z_{q}^{2}, \ldots, z_{q}^{m})\). The loss function for this dynamic quantization-based VQ-VAE model is given in Equation \ref{eq4}.
\begin{equation}
\label{eq4}
\mathcal{L} =\left \| x-\hat{x}  \right \| _{2}^{2} +\frac{1}{m} \sum_{i=1}^{m} \beta \left \{  ||sg[z_{e}]-z_{q}^{i}  ||_{2}^{2}+\alpha||z_{e}-sg[z_{q}^{i}]||_{2}^{2}]\right \} 
\end{equation}
\section{Experimental results and analysis}
\label{sec4}
All our experiments were conducted with a constant information capacity for the codebooks. This means that while the sizes of the codebooks and the embedding dimensions varied, their product remained constant, ensuring the discrete space capacity stayed unchanged. We experimented with six datasets: MNIST, FashionMNIST, CIFAR-10, Tiny-Imagenet, CelebA, and a real-world dataset, the Diabetic Retinopathy dataset. In the experiments described in Sections \ref{sec4.1} and \ref{sec4.2}, the discrete information space size of the experimental model was set to \(65536\). The multi-head attention mechanism used had two heads. The codebook in the VQ operation was updated using the EMA method. Both the encoder and decoder architectures were designed with Convolutional Neural Networks(CNN)\citep{lecun1998gradient} and Residual Network(ResNet)\citep{he2016deep}. The hyperparameters \(\alpha\), \(\beta\), and \(\gamma\) were set to \(0.25\), \(1\), and \(0.99\), respectively. Part of the experimental results are presented in Section \ref{sec4}, please refer to the \nameref{appendix} for other results.

\subsection{The influence of codebook size and embedding dimension}\label{sec4.1}
\begin{figure}[h]
  \centering
  \begin{subfigure}[b]{0.8\linewidth}
    \raggedright
    \resizebox{\linewidth}{!}{\includegraphics{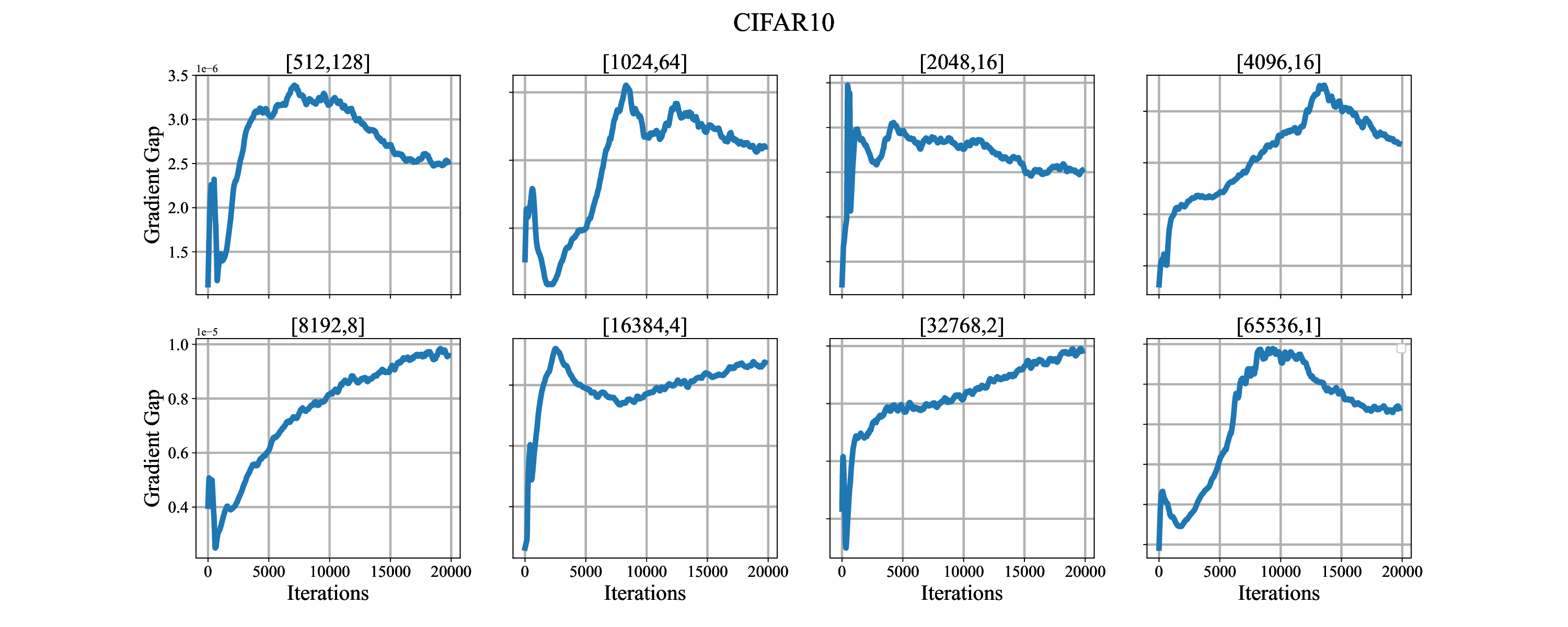}}
  \end{subfigure}%
  \hspace{-0.9cm}
  \begin{subfigure}[b]{0.8\linewidth}
    \raggedright
    \resizebox{\linewidth}{!}{\includegraphics{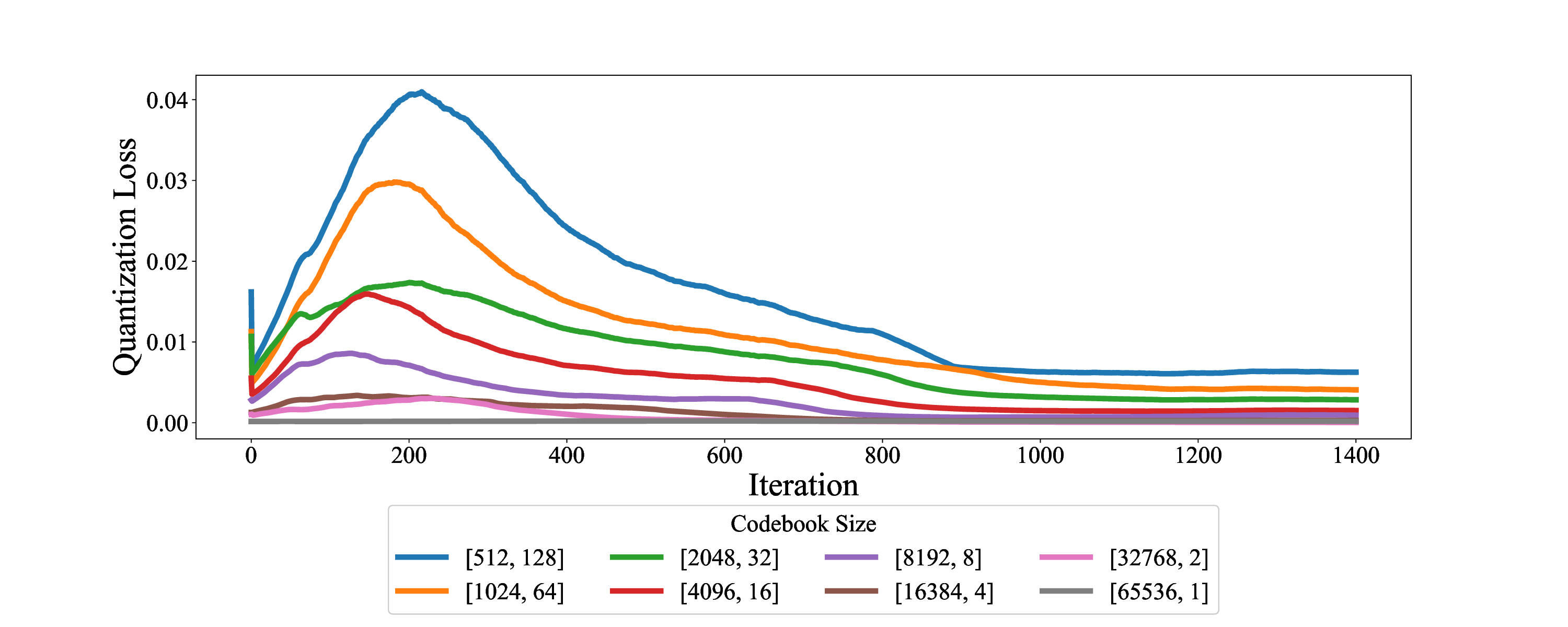}}
  \end{subfigure}
  \caption{Gradient gap and quantization loss under fixed codebook models.}
  \label{fig2}
\end{figure}

Before enhancing the quantization method of the VQ-VAE, we conducted an in-depth analysis on how various combinations of codebook size \(N\) and embedding dimensions \(D\) affect the model’s performance. This study was carried out with the constraint that the capacity of the discrete space in the VQ-VAE model remained constant, meaning that the product of \(N\) and \(D\) was kept unchanged. In the original VQ-VAE model configuration, we specified a codebook size of \(512\) and embedding dimensions of \(128\). From this starting point, we transitioned to experimenting with various other combinations of \(N\) (codebook size) and \(D\) (embedding dimensions), ranging from \([1024, 64]\) to \([65536, 1]\), to observe their individual effects on the model's performance. Our findings indicate that increasing \(N\), the codebook size, generally enhances the model's performance. However, maintaining a sufficiently high \(D\), the embedding dimension, is also essential to sustain this improvement. The results from these investigations are displayed in Table~\ref{table1}, which outlines the reconstruction loss on the validation set for various \(N, D\) configurations, each maintained under identical training conditions across six datasets.
\begin{table}
  \caption{Reconstruction loss under different fixed codebook models}
  \centering
  \setlength{\tabcolsep}{2pt}
  \begin{tabular}{ccccccc}
    \toprule
    \multicolumn{7}{c}{Datasets}
    \\
    \cmidrule(r){2-7}
    [N,D] & MNIST & FshionMNIST & CIFAR10 & Tiny-ImageNet  & Diabetic-retinopathy & CelebA\\
    \midrule
    {[512,128]}           & 0.285                 & 0.577           & 0.606     & 2.679   & 0.361 & 0.177 \\
    {[1024,64]}         & 0.294                  & 0.590          & 0.663     &  2.654 & 0.400  & 0.156\\
    {[2048,32]}          & 0.291                  & 0.602           & 0.618     & 2.640 & 0.410 & 0.155 \\
    {[4096,16]}         & 0.265                  & 0.589           & 0.618     & 2.650 & 0.362 & 0.136\\
    {[8192,8]}         & 0.240                  & 0.503           & 0.553     & 2.605  & \textbf{0.337} & 0.122\\
    {[16384,4]}         & 0.210                  & \textbf{0.489}          & \textbf{0.503}     & \textbf{2.528}  & 0.347 & \textbf{0.120} \\
    {[32768,2]}     & \textbf{0.199}  & 0.578           & 0.808     & 3.364 & 0.407  & 0.171 \\
    {[65536,1]}         & 0.377                  & 0.877           & 1.353     & 4.631 & 0.493  & 0.334 \\
    \bottomrule
  \end{tabular}
  \label{table1}
\end{table}
Table \ref{table1} illustrates that smaller embedding dimensions in the codebook do not always lead to better results. In the original VQ-VAE model, while the codebook size remains constant, modifying it and retraining can yield varied outcomes. Furthermore, within a fixed discrete information space, there is an optimal codebook size that best fits the specific dataset. Following these findings, we computed the terminal gradient gap\citep{huh2023straightening} for models that use a consistent codebook size throughout their training. A successful quantization function \(h\left ( \cdot \right )\) retains essential information from a finite vector set \(z_{e}\). The quantized vector is expressed as \(z_{q} = z_{e} + \varepsilon\), where \(\varepsilon\) is the residual vector derived from the quantization process. If \(\varepsilon = 0\), it implies no direct estimation error, suggesting that the model effectively lacks a quantization function. Equation \ref{eq5} describes how to measure the gradient gap concerning this ideal, lossless quantization function.
\begin{equation}
\label{eq5}
\Delta{Gap}=\left\|\frac{\partial\mathcal{L}_{\text{task}}(G(\mathbf{z}_e))}{\partial E(x)}-\frac{\partial\mathcal{L}_{\text{task}}(G(\mathbf{z}_q))}{\partial E(x)}\right\|
\end{equation}
During the training phase, we monitored the variations in the gradient gap. Notably, the pattern of gradient gap changes associated with the codebook remained consistent across three-channel datasets. For instance, we demonstrated the evolving pattern of the gradient gap for the CIFAR10 dataset in Figure \ref{fig2}(left). Although the gradient gap changes are quite similar between two single-channel datasets, differences emerge when comparing single-channel and three-channel datasets. More comprehensive details on these differences are discussed in the \nameref{appendix}.
The reconstruction loss for the other three datasets shows a similar trend. As the codebook size increases up to the point of achieving the lowest validation loss, the gradient gap remains constant. However, once the size surpasses this optimal codebook combination, any further decrease in embedding dimension results in a progressive increase in the gradient gap for subsequent models. From the findings in Table \ref{table1}, we conclude that within a fixed discrete information space, enlarging the codebook size to a certain extent can enhance the model's reconstruction capabilities. Nonetheless, if the embedding dimension becomes too small, it will lead to an increased quantization gradient gap, which could degrade the model's performance rather than enhance it.

Given that each batch contains an equal number of data points, the quantization loss is impacted by the embedding dimension. A smaller embedding dimension tends to yield a reduced quantization loss, as depicted in Figure \ref{fig2}(right).

\textbf{Analysis}: Increasing \(N\) generally allows the model to capture more variability in the data due to a larger number of vectors available for encoding the data, potentially reducing quantization error. However, as \(N\) increases, the dimension \(D\) typically decreases to manage computational complexity and memory usage, which can affect the quality of each individual representation. Lower \(D\) might lead to loss of information, since each vector has less capacity to encapsulate details of the data. However, too high a dimension could also mean that the vectors are too specific, reducing their generalizability.  As \(N\) increases, each vector in the codebook needs to handle less variance, theoretically reducing the reconstruction loss. However, each vector also has fewer \(D\) to capture that variance as \(N\) decreases. The balance between these factors is delicate and dataset-specific. The characteristics of each dataset heavily influence the optimal \(N\) and \(D\) . Simpler datasets with less intrinsic variability (like MNIST) can be effectively modeled even with lower \(D\) , while more complex datasets require either higher \(D\) or more carefully balanced \(D\).
\subsection{Adaptive dynamic quantization mechanism}
\label{sec4.2}
The adaptive dynamic quantization mechanism allows each data point in a batch to independently select the most appropriate codebook for quantization and training. Mirroring the format seen in Table \ref{table1}, we have showcased in Table \ref{table2} the comparison of reconstruction loss between the best fixed codebook model an adaptive dynamic quantization model across various validation datasets.
\begin{table}
  \caption{Comparison of reconstruction performance between adaptive dynamic quantization model and optimal fixed codebook model.}
  \centering
  \setlength{\tabcolsep}{2pt}
  \begin{tabular}{ccccccc}
    \toprule
    \multicolumn{7}{c}{Datasets}
    \\
    \cmidrule(r){2-7}
    Models &  MNIST  & FshionMNIST & CIFAR10 & Tiny-ImageNet & Diabetic-retinopathy  & CelebA\\  
    \midrule
	Fixed Codebook           & 0.199  & 0.489  & 0.503  & 2.528 & 0.337  & 0.120  \\  
    Adaptive Codebook         & \textbf{0.166}  & \textbf{0.410}  & \textbf{0.394}  & \textbf{2.022} & \textbf{0.294} & \textbf{0.099} \\ 
    \bottomrule
  \end{tabular}
  \label{table2}
\end{table}
Even with an optimized fixed quantization codebook for reconstruction performance, it falls short compared to our proposed adaptive dynamic quantization model. This is because our model allows each input data point to choose from eight codebooks within a discrete information space of 65536 possibilities, whereas the fixed codebook model offers no such flexibility.

\begin{figure}[htbp]
    \centering
    \begin{minipage}[t]{0.49\textwidth}
        \centering
        \includegraphics[width=\textwidth]{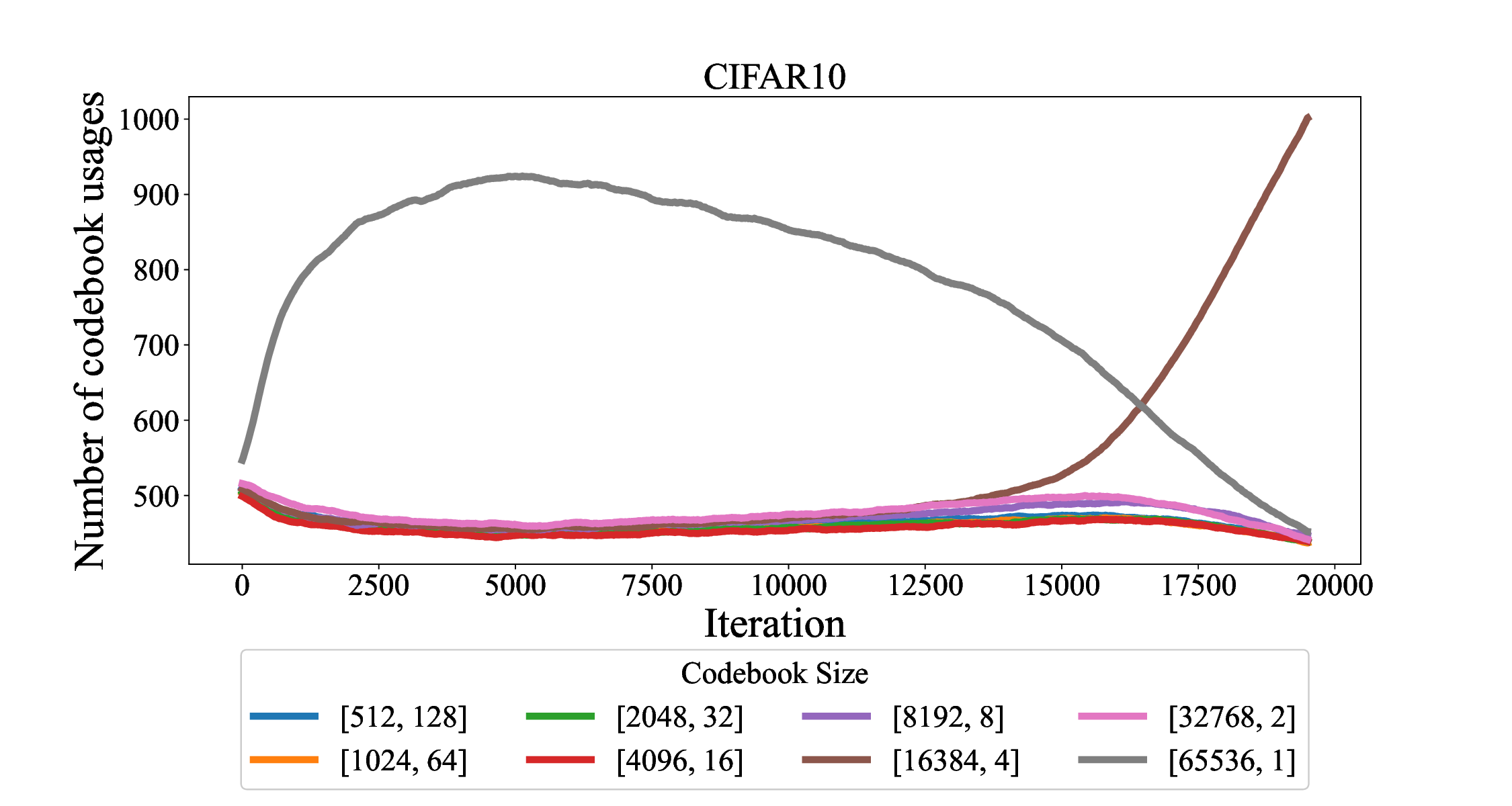}
    \end{minipage}
    \hfill
    \begin{minipage}[t]{0.49\textwidth}
        \centering
        \includegraphics[width=\textwidth]{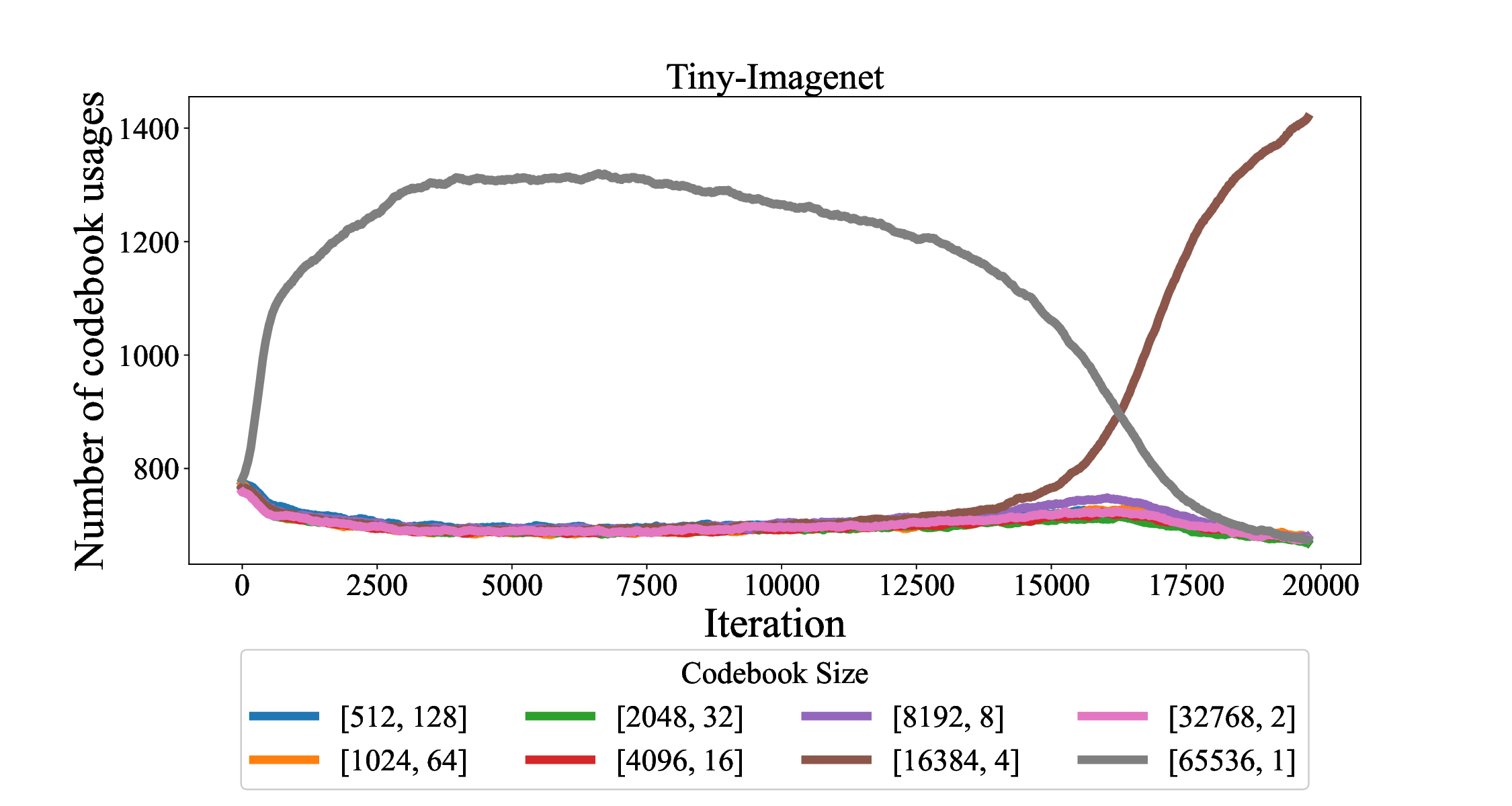}
    \end{minipage}
    \hfill
    \begin{minipage}[t]{0.49\textwidth}
        \centering
        \includegraphics[width=\textwidth]{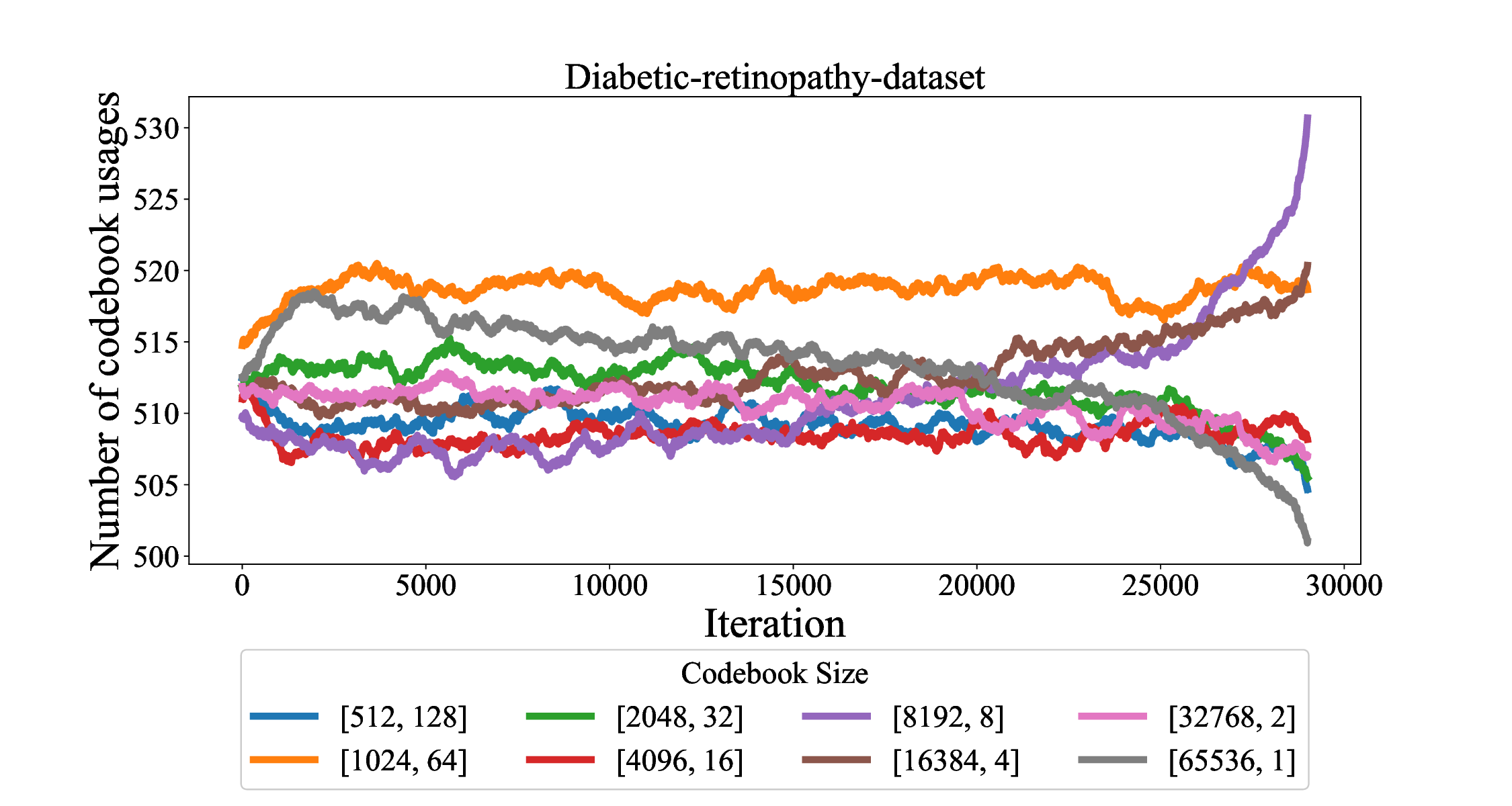}
    \end{minipage}
    \begin{minipage}[t]{0.49\textwidth}
        \centering
        \includegraphics[width=\textwidth]{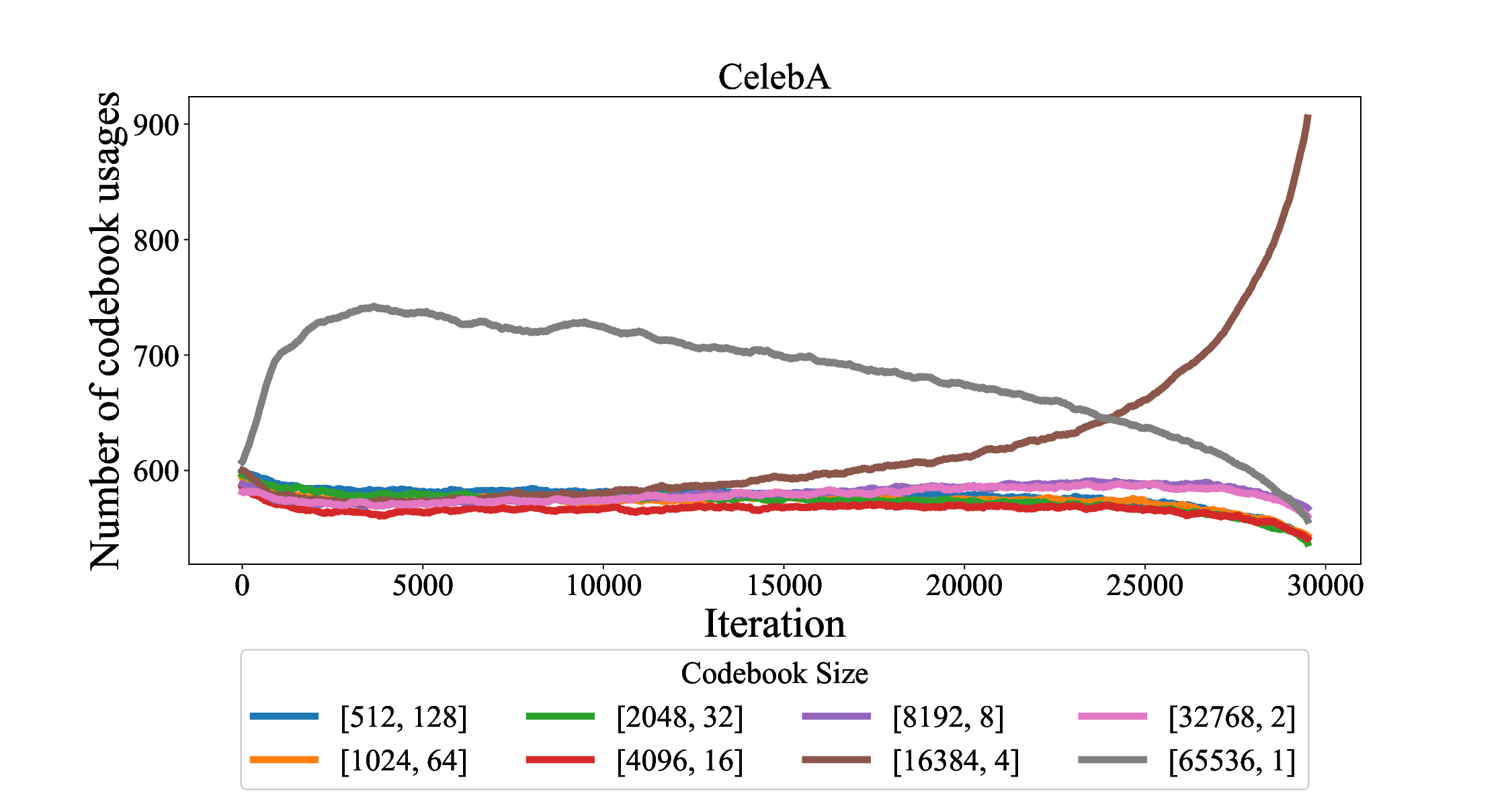}
    \end{minipage}
    \caption{The frequency of adaptive selection of different-sized codebooks throughout the training process. The training can be divided into two stages: Initially, the model predominantly learns from the codebook with the most codewords; Subsequently, the model optimizes and increasingly selects the type of codebook with the minimum reconstruction loss in the fixed codebook model. The experimental results on the Diabetic Retinopathy dataset are slightly less conspicuous, but in the final stage of training, there is still a tendency to choose the most suitable codebook}
    \label{fig3}
\end{figure}

The main advantage of our method lies in its ability to learn the distribution of different features in the dataset and adapt to the codebook during training. To illustrate the usage of different codebooks, we plotted the frequency of codebook usage in the adaptive dynamic quantization model throughout the training process. The results align closely with the data in Section \ref{sec4.1}. As shown in Figure \ref{fig3}, the adaptive dynamic quantization model, implemented with Gumbel-softmax and an attention mechanism, gradually learns the various feature distributions of the dataset and selects different codebooks. At the beginning, the model tends to choose the largest codebook for fast learning, while in the later stage of learning, it tends to choose the most suitable codebook. By the final stage of training, the codebook size with the highest usage frequency matches the codebook size that achieved the lowest reconstruction loss in the fixed codebook experiment detailed in Section \ref{sec4.1}. For the Diabetic Retinopathy dataset, our analysis is that the dataset itself has not been uniformly preprocessed and the distribution of different features in the dataset is not obvious(As shown in Figure \ref{fig5}), resulting in a lack of inclination to choose a certain codebook at the beginning of training. However, in the final stage, the model will gradually converge and tend to choose the most suitable codebook.

\begin{figure}[h]
  \centering
    \begin{minipage}[t]{0.49\textwidth}
        \includegraphics[width=\textwidth]{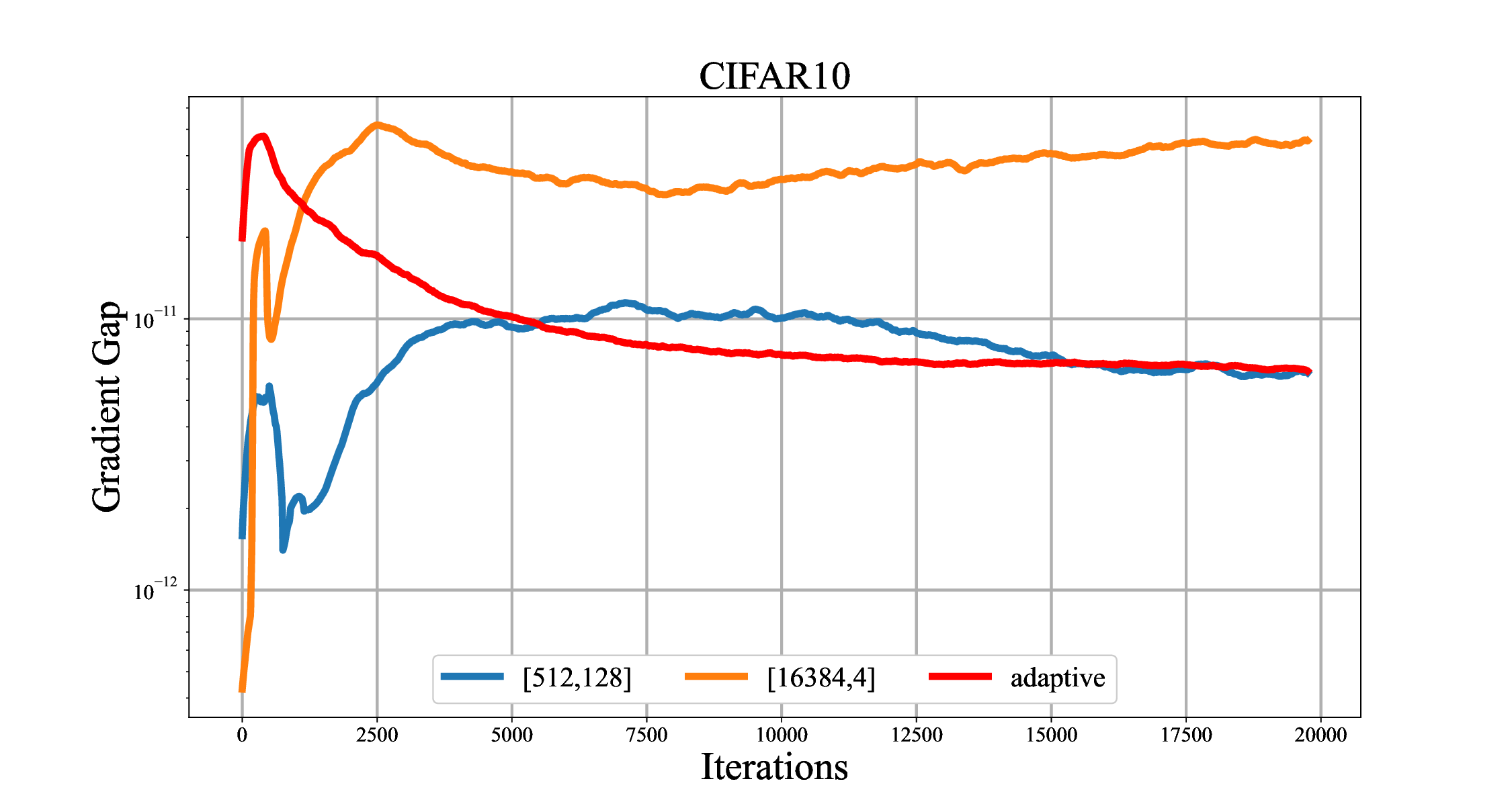}
    \end{minipage}
    \begin{minipage}[t]{0.49\textwidth}
        \includegraphics[width=\textwidth]{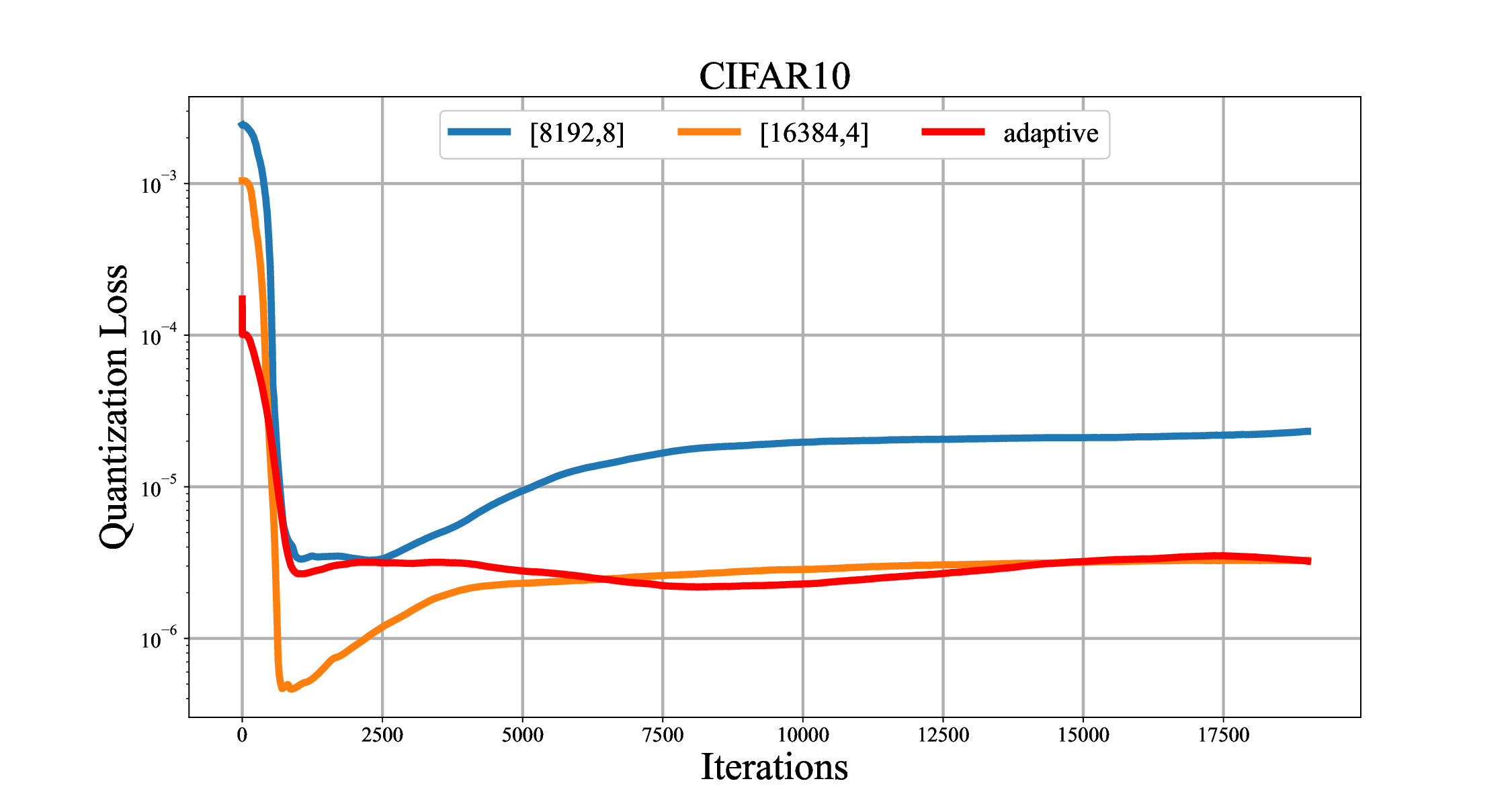}
    \end{minipage}
    \hfill
    \begin{minipage}[t]{0.49\textwidth}
        \includegraphics[width=\textwidth]{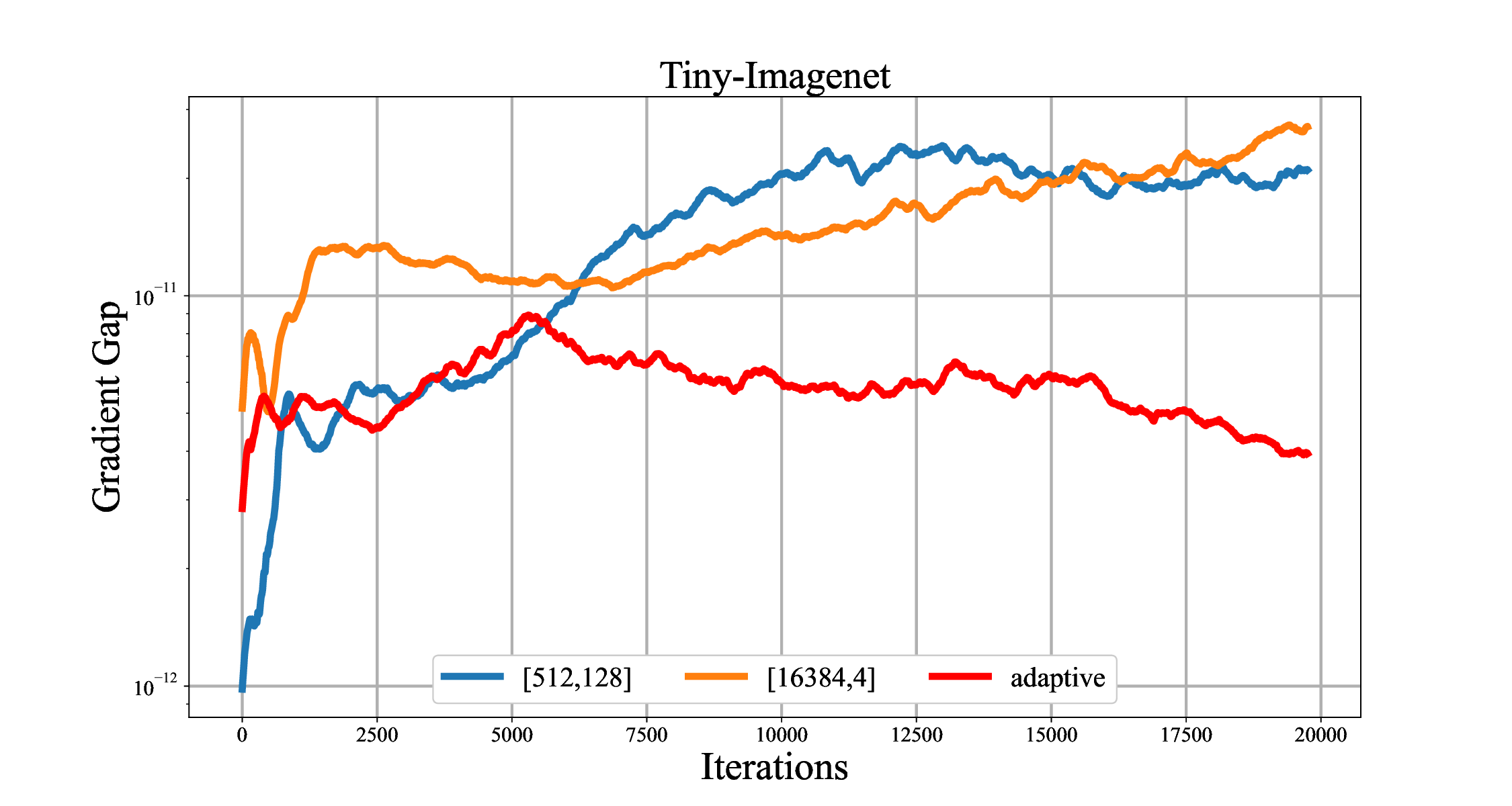}
    \end{minipage}
    \begin{minipage}[t]{0.49\textwidth}
        \includegraphics[width=\textwidth]{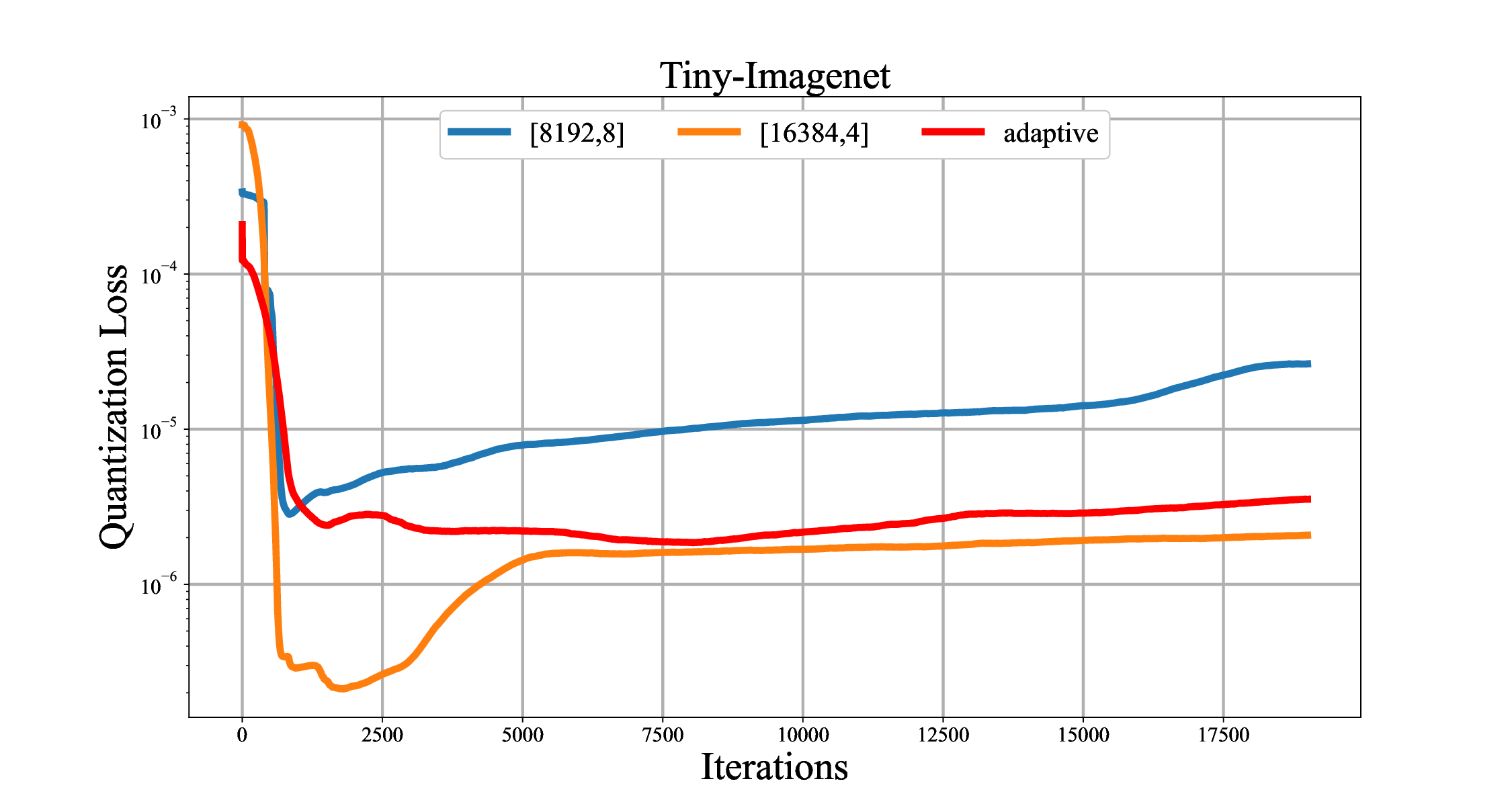}
    \end{minipage}
    \hfill
    \begin{minipage}[t]{0.49\textwidth}
        \includegraphics[width=\textwidth]{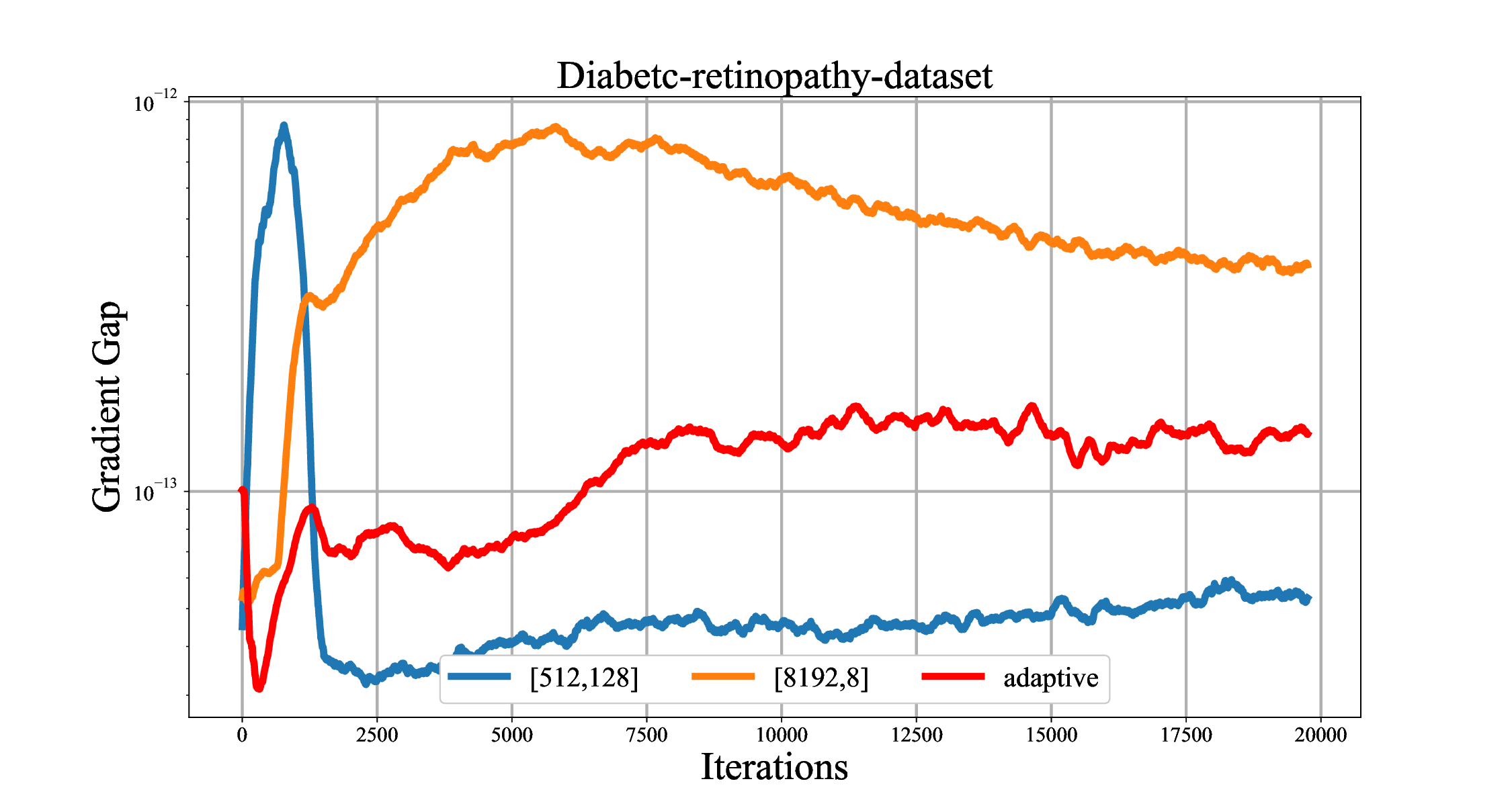}
    \end{minipage}\vspace{-0.1cm}
    \begin{minipage}[t]{0.49\textwidth}
        \includegraphics[width=\textwidth]{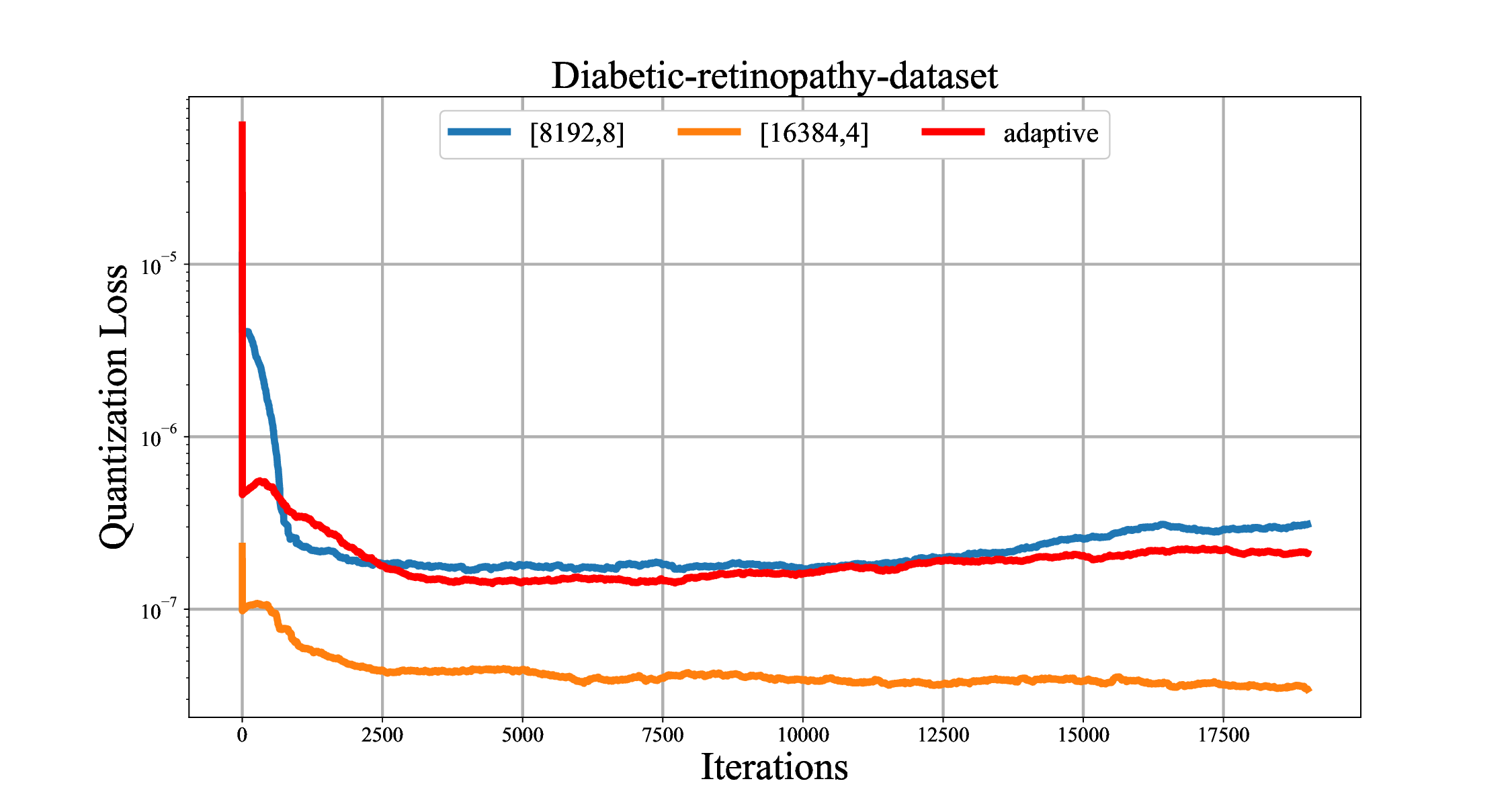}
    \end{minipage}
    \hfill
    \begin{minipage}[t]{0.49\textwidth}
        \includegraphics[width=\textwidth]{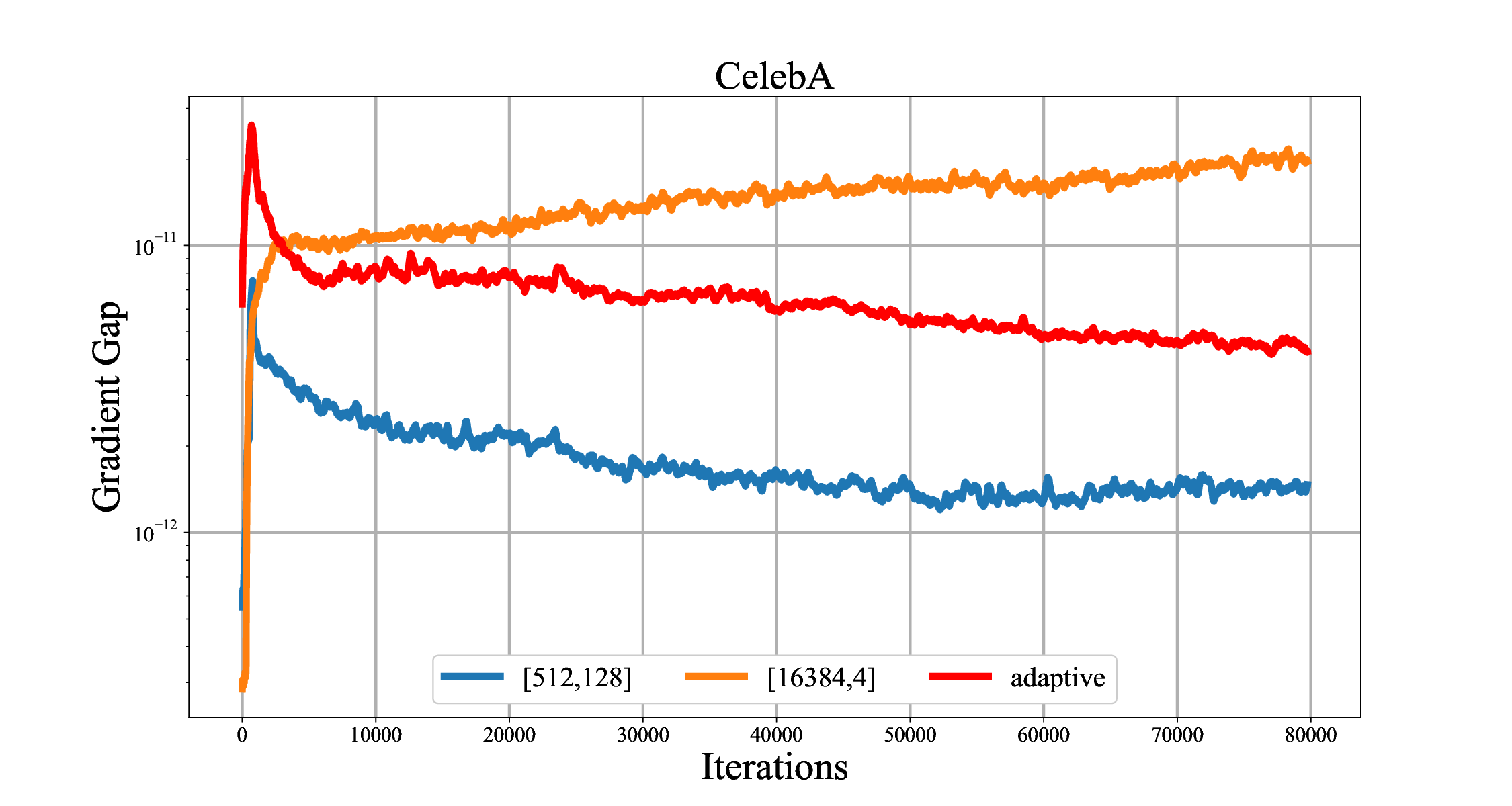}
    \end{minipage}\vspace{-0.1cm}
    \begin{minipage}[t]{0.49\textwidth}
        \includegraphics[width=\textwidth]{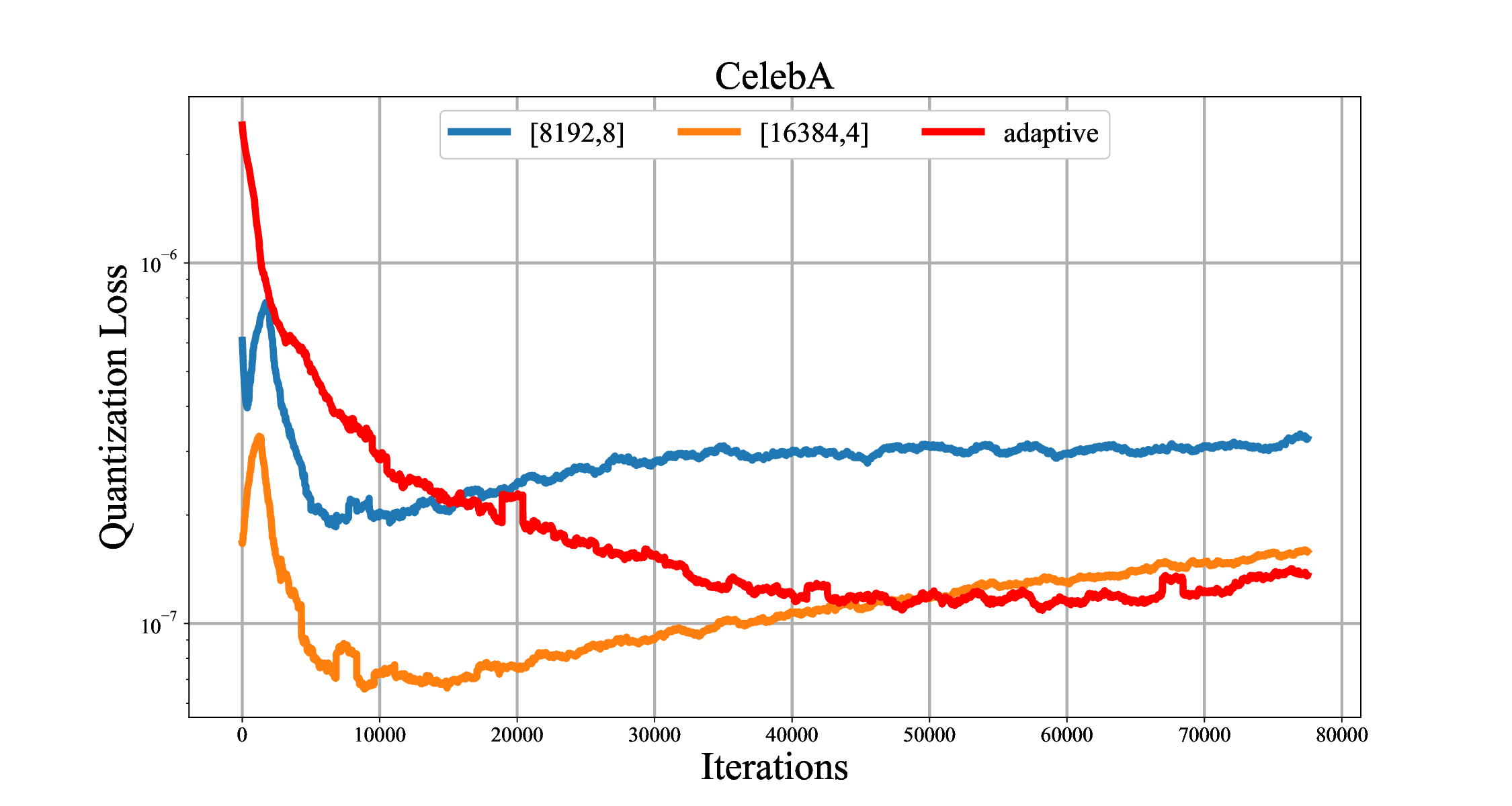}
    \end{minipage}\vspace{-0.1cm}
  \caption{
Comparison of quantization loss and gradient gap between adaptive dynamic quantization model and fixed codebook model. The blue color represents the model with the smallest gradient gap and the largest quantization loss in the fixed codebook model, the yellow color represents the fixed codebook model with the smallest reconstruction loss, and the red color represents the adaptive dynamic quantization model. The gradient gap and quantization loss of the adaptive quantizer are kept at low values.}
  \label{fig4}
\end{figure}

We plotteed a comparison graph of the gradient gap produced by our method versus that produced by fixed codebooks. The results, displayed in Figure \ref{fig4}(left), show that the adaptive dynamic quantizer method achieves optimal reconstruction performance while maintaining low gradient gap. The size of the gradient difference depends on the magnitude of the quantization error and the smoothness of the decoding function \( G\left( \cdot \right) \). The minimal change in quantization loss indicates that the decoder of the adaptive dynamic quantization model exhibits better smoothness compared to the fixed codebook model. 
The pattern in Figure \ref{fig2}(right) is clear: larger codebook sizes correspond to smaller embedding dimensions and lower quantization losses. We compared the quantization loss of the dynamic quantizer with various fixed codebooks and plotted the range of the dynamic quantizer's quantization loss. The quantization loss resulting from our dynamic quantization method is close to that of the optimal fixed codebook, demonstrating the effectiveness of our approach. As shown in Figure \ref{fig4}(right), the y-axis is represented in logarithmic form for clarity. 

\textbf{Analysis}: At the beginning of training, the model is in an exploratory phase. It is trying to learn and understand the structure of the data. Larger codebook size provide a finer granularity of representation and offer more codewords to choose from, which allows the model to capture more detailed and diverse patterns in the data. Using larger codebook size can help the model avoid early commitment to suboptimal quantization, giving it the flexibility to explore different parts of the data space. As training progresses, the model starts to specialise and optimise its parameters based on the data it has seen. The need for the fine granularity provided by the larger codebook size diminishes because the model becomes more confident in its learned representations. They help prevent overfitting by providing a more generalised representation of the data. With fewer codewords, the model is forced to generalise better, which can lead to improved performance on unseen data. Gradually transitioning to the most suitable codebook size helps strike a balance between accurate data representation and model generalisation. For detailed mathematical proofs, please refer to the \nameref{appendix}.

\subsection{Albation study}
\begin{table}
  \caption{Comparison of reconstruction performance among variants of the model that adaptively select codebooks.}
  \centering
  \setlength{\tabcolsep}{2pt}
  \begin{tabular}{ccccccc}
    \toprule
    \multicolumn{7}{c}{Datasets}
    \\
    \cmidrule(r){2-7}
    Models & MNIST & FshionMNIST & CIFAR10 & Tiny-ImageNet & Diabetic-retinopathy  & CelebA\\
    \midrule
    Base           & \textbf{0.165} 
                 & \textbf{0.410} 
           & \textbf{0.394} 
     & \textbf{2.022}    & \textbf{0.294}  & \textbf{0.099}\\
    {\(W=32786\)}         & 0.176                  & 0.436          & 0.412     &  2.156         &  0.338 &  0.106\\
    {\(W=16384\)}          & 0.181                  & 0.442           & 0.427     & 2.207        & 0.337 &  0.108\\
    {\(W=8192\)}         & 0.191                  & 0.447           & 0.427     & 2.329          & 0.305  &  0.104\\
    {\(W=4096\)}         & 0.193                  & 0.447          & 0.458     & 2.336  & 0.358   &  0.104 \\
    {Without EMA}         & 0.317                  & 0.595          & 0.490     & 2.559   & 0.361  &  0.173 \\
    {\(\alpha=0.5\)}     & 0.171  & 0.424          & 0.428     & 2.204  & 0.355  &  0.102 \\
    {\(\alpha=0.75\)}         & 0.175                 & 0.442           & 0.440     & 2.482   & 0.325 &  0.104 \\
    {\(\alpha=1.0\)}          & 0.188                 & 0.492           & 0.466     & 2.387        & 0.380 &  0.118 \\
    {\(\alpha=5\)}         & 0.257                  & 0.682           & 0.630     & 3.027         & 0.464 &  0.138 \\
    {\(\alpha=10\)}         & 0.271                  & 0.823          & 0.700     & 3.278  & 0.485  &  0.182 \\
    {\(\beta=0.01\)}         & 0.169                  & 0.427          & 0.381     & 2.179   & 0.299 &  0.091  \\
    {\(\beta=0.2\)}     & 0.167  & \textbf{0.397}          & \textbf{0.349}   & \textbf{1.999}  & \textbf{0.279}  & \textbf{0.088}\\
    {\(\beta=5\)}         & 0.191                & 0.487           & 0.491    & 2.462   & 0.360 & 0.110 \\
    {CNN}     & \textbf{0.154}  & \textbf{0.365}          & \textbf{0.330}     & \textbf{1.828}  & \textbf{0.293} & \textbf{0.091} \\
    {CNN+Inception}         & 0.190                 & 0.539          & 0.459     & 2.467   & 0.360 & 0.104\\
    \bottomrule
  \end{tabular}
  \label{table3}
\end{table}

The 'Base' model in Table \ref{table3} refers to the model discussed at the beginning of Section \ref{sec4}. Building on this model configuration, we conducted ablation experiments addressing various aspects such as the model's encoder-decoder structure, hyperparameters, discrete information space, and the use of EMA. Increasing the capacity of the discrete space and employing EMA to update codebooks both enhance the model's performance, with the base experiment's value of \(\alpha=0.25\) being appropriate. Across these datasets, reducing the proportion of quantization loss (\(\beta=0.2\)) is shown to be beneficial. Additionally, it is noteworthy that these datasets do not require overly complex encoders and decoders. Adding ResNet and Inception layers \citep{szegedy2015going} after the convolutional layers actually decreases model performance. The loss values in Tables \ref{table1}, \ref{table2}, and \ref{table3} represent the sum of reconstruction losses incurred by the model when evaluated on the respective standard validation sets of the corresponding datasets after one complete pass.



\section{Conclusion}\label{sec5}
In this study, we focused on the impact of the number of codebooks and the embedding dimension on VQ-VAE's performance, with particular attention to the situation where the product of these two parameters remains constant. Our research introduced an adaptive dynamic quantizer based on Gumbel-softmax, allowing the model to adaptively choose the most ideal codebook size and embedding dimension at any given data point. It was found that the model, adopting the new method, is divided into two stages throughout the learning process. In the first stage, it tends to learn using codebooks of the largest size, while in the second stage, it gradually favors the use of the most suitable codebook. Our method demonstrated significant performance improvements in empirical validation across multiple benchmark datasets, proving the importance of adaptive dynamic discretizers for optimizing VQ-VAE models.
\newpage
\bibliographystyle{plainnat} 
\bibliography{refer} 

\newpage
\appendix
\section{Appendix}\label{appendix}
\subsection{Experimental details}\label{app1}
\begin{figure}[htbp]
	\begin{minipage}{0.5\linewidth}
		\centering
		\includegraphics[width=0.8\linewidth]{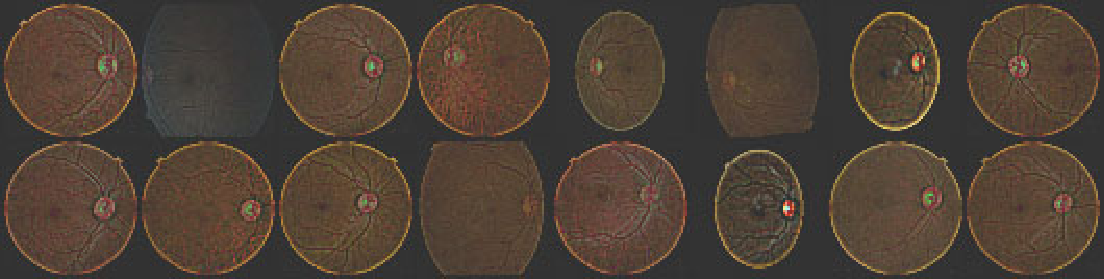}
	\end{minipage}
        \begin{minipage}{0.5\linewidth}
		\centering
		\includegraphics[width=0.8\linewidth]{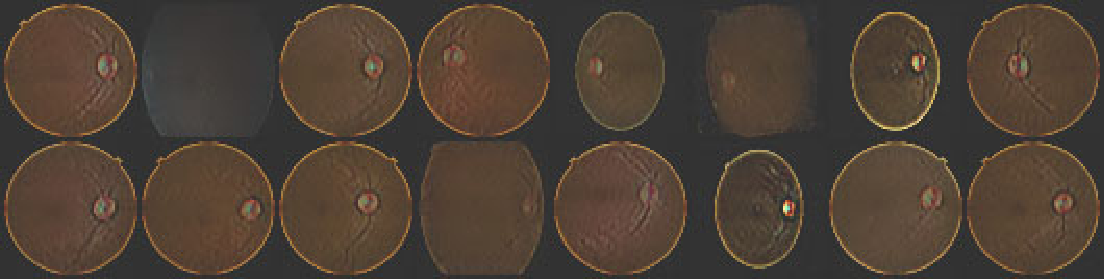}
	\end{minipage}
        \caption{Sample images of Diabetic Retinopathy dataset. Left: Original image. Right: Reconstructed image.}
        \label{fig5}
\end{figure}

Diabetic Retinopathy dataset consists of a total of \(2750\) images belonging to five different categories of symptoms. All images have a size of \(256\times 256\) pixels. During the experiments, we selected \(250\) images as a validation set, following a proportional distribution among the five categories. The remaining \(2500\) images were used as the training set. Figure \ref{fig5} shows some example images from this dataset. The CelebA dataset we used in all experiments was \(8000\) images extracted from official sources, extract \(1000\) images from each of the eight attributes, with \(920\) images from each category as the training set and \(80\) images as the validation set. Regarding training, due to the limitation of computer memory, the batch sizes for MNIST, FashionMNIST, and CIFAR10 are all set to \(64\). For Tiny-imagenet, CelebA and Diabetic-retinopathy-dataset, the batch sizes are set to \(24\), \(2\) and \(1\) respectively. The learning rate for model training in the experiment is set to \(1e-4\), and the Adam optimizer is used. The temperature in the Gumbel-softmax function decreases as training progresses and kept at \(1\) during validation. Assuming \(m\) represents the \(m\)-th batch of training that the model is currently undergoing, then \(Temperature_{training} = \left (  iterations-m\right ) +  1,Temperature_{validation} =1\).
\subsection{Result analysis and proof}\label{app2}
\begin{figure}[htbp]
    \centering
    \begin{minipage}[t]{0.49\textwidth}
        \centering
        \includegraphics[width=\textwidth]{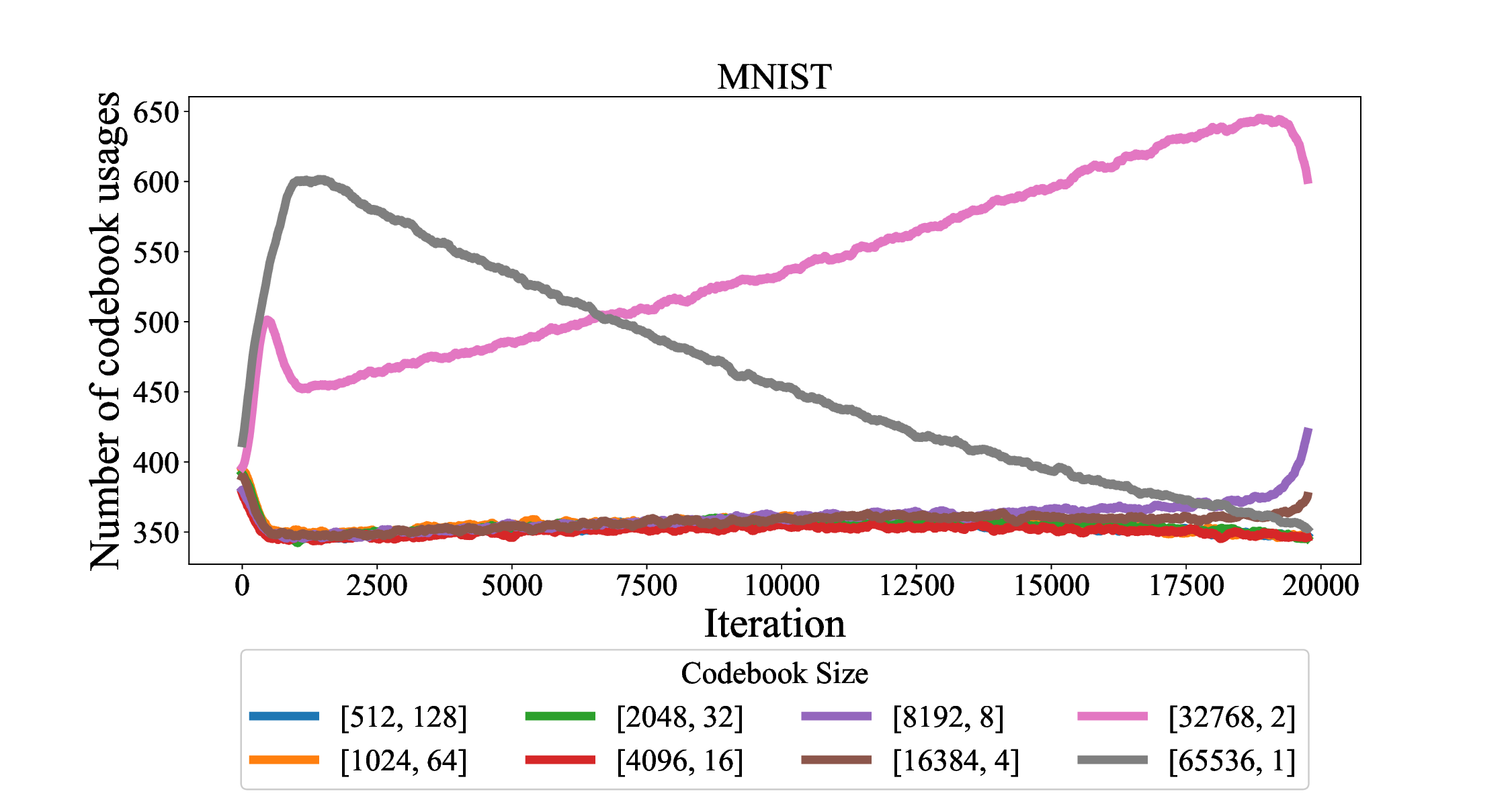}
    \end{minipage}
    \hfill
    \begin{minipage}[t]{0.49\textwidth}
        \centering
        \includegraphics[width=\textwidth]{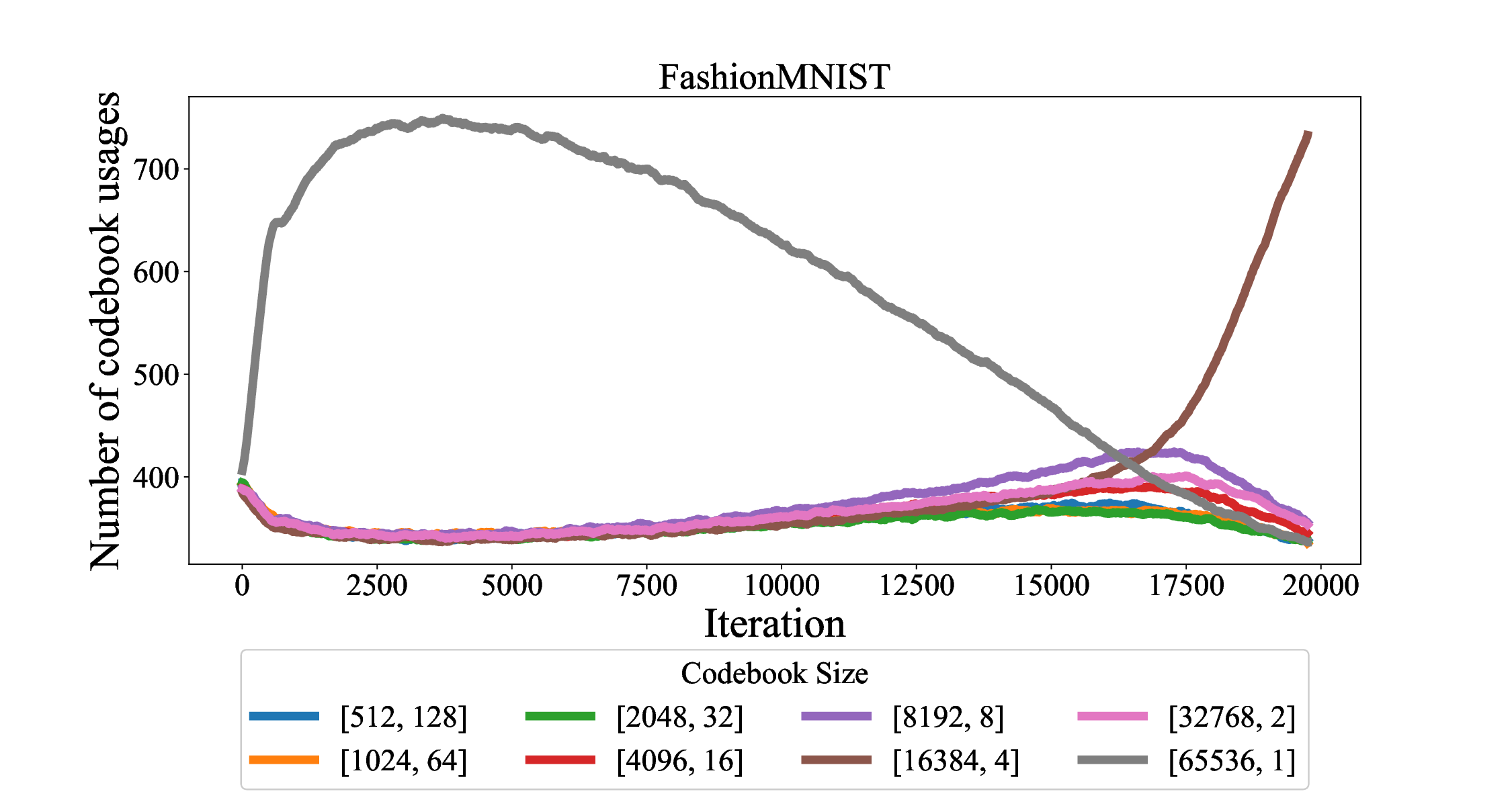}
    \end{minipage}
    \caption{The frequency of selecting different codebooks during training of the adaptive dynamic quantization model on single channel datasets. Our experimental results are very consistent across five official datasets, where the adaptive codebook quantizer always manages to find the most suitable fixed codebook for its dataset at the last stage of learning. }
    \label{fig6}
\end{figure}

In Figure \ref{fig1}, we clearly demonstrate how to utilize the multi-head attention mechanism and the Gumbel-Softmax method to implement the dynamic selection of quantization codebooks in the model. Of course, the candidate codebook items are limited, as having too many quantization codebooks would make the model complex and consume computational resources. Figure \ref{fig6} shows the experimental results on two single channel datasets. 
Here, we provide our analysis based on our experimental results. 

\textbf{Reconstruction Loss Definition}: Given a dataset \(\mathcal{X}=\{\mathbf{x}_i\}_{i=1}^n\mathrm{~where~}\mathbf{x}_i\in\mathbb{R}^M\),  the reconstruction loss \(L\) using vector quantization is defined as:
\begin{equation}
\label{eq6}
L=\frac{1}{n}\sum_{i=1}^n\|\mathbf{x}_i-\mathbf{\hat{x}}_i\|^2
\end{equation}
where \(\hat{x}_i\) is the nearest codebook vector to \({x}_i\).

\textbf{Representational Capacity}: The representational capacity of a codebook depends on its size \(N\) and dimensionality \(D\). We introduce a constant \(K\) representing the inherent complexity of the data:
\begin{equation}
\label{eq7}
C\propto K\cdot N\cdot D
\end{equation}
where \(K\) is proportional to the data’s intrinsic dimensionality and variability.

\textbf{Quantization Error with Large Codebook Size}: As the codebook size \(N\) increases, the model starts to memorize the training data. This memorization can be quantified by the variance in the reconstruction error. Let \(V\) represent the variance of the data:
\begin{equation}
\label{eq8}
Quantization\quad Error \propto \frac{V}{N} 
\end{equation}

\textbf{Representation Error Due to Reduced Dimensionality
}: When \(N\) increases, \(D\) often decreases to keep the overall complexity manageable.
As \(D\) decreases, the ability of the codebook vectors to represent the data reduces. The reconstruction error due to reduced dimensionality can be expressed with a constant \(\alpha\), reflecting the complexity of the data:
\begin{equation}
\label{eq9}
Representation\quad Error \propto \frac{\alpha }{D} 
\end{equation}
Assuming \(D=\frac{\beta }{N}\) (where \(\beta\) is a constant related to the total codebook capacity), we get:
\begin{equation}
\label{eq10}
Representation\quad Error \propto \frac{\alpha N }{\beta } 
\end{equation}

\textbf{Total Reconstruction Loss}: The total reconstruction loss \(L\) is a combination of the errors due to quantization
and reduced representation:
\begin{equation}
\label{eq11}
L\propto \frac{V}{N} +\frac{\alpha N}{\beta } 
\end{equation}

\textbf{Incorporating Constants and Minimizing Loss}: Let’s incorporate the constants \(K\), \(V\), \(\alpha\), and \(\beta\):
\begin{equation}
\label{eq12}
L=\frac{V}{N} +\frac{\alpha kN}{\beta } 
\end{equation}
To find the optimal codebook size \(N_{optimal}\), we take the derivative of \(L\) with respect to \(N\) and set it to \(0\):
\begin{equation}
\label{eq13}
\frac{dL}{dN}=-\frac{V}{N^2}+\frac{\alpha K}{\beta}=0
\end{equation}
Solving for \(N\), we get:
\begin{equation}
\label{eq14}
\frac{V}{N^2}=\frac{\alpha K}{\beta}
\end{equation}
\begin{equation}
\label{eq15}
N_{optimal} =\sqrt{\frac{V\beta }{\alpha K} } 
\end{equation}

\textbf{Behavior Beyond the Optimal Point}: For \(N\) beyond \(N_{optimal}\): 
\begin{equation}
\label{eq16}
N_{optimal} =\sqrt{\frac{V\beta }{\alpha K} } 
\end{equation}
When \(N > N_{optimal}\), the term \(\frac{\alpha K N}{\beta}\)
dominates, leading to an increase in reconstruction loss \(L\). Thus:
\begin{equation}
\label{eq17}
L\propto \frac{\alpha K N}{\beta}
\end{equation}
This refined proof demonstrates that the reconstruction loss \(L\) initially decreases with increasing codebook size \(N\) due to improved data representation. However, beyond a certain point, further increasing \(N\) leads to higher loss due to:
\begin{itemize}
\item \textbf{Quantization Error:} Captured by the term \(\frac{V}{N}\), indicating that as \(N\) increases, the model becomes too specific to the training data, increasing reconstruction loss on new data.
\item \textbf{Representation Error:}  Captured by the term \(\frac{\alpha K N}{\beta}\), indicating that as \(D\) (which is inversely proportional to \(N\)) decreases, the codebook vectors cannot adequately capture the data complexity, leading to increased reconstruction loss.
\end{itemize}
\subsection{Limitation}\label{app3}
\begin{itemize}
\item \textbf{Research on Generative Models:} Our research method mainly focuses on exploring and optimizing reconstruction performance. Although significant achievements have been made in exploring this direction, we must acknowledge that due to the concentration of research objectives, we have not conducted equally in-depth research on optimizing generation performance. This is a research direction for our future work. 
\item \textbf{Finite Discrete Codebook Space:} We conducted our research under limited computing resources. Further exploration is needed to determine what happens when the discrete codebook space is larger and there are more codebook options available. Perhaps there will be more training stages and more optimal selection options
\end{itemize}
\subsection{Additional supplementary materials for experiments}\label{app4}
Equation \ref{eq5} mentions that gradient gap measures the gradient difference between the non-quantized model and the quantized model. When the gap is \(0\), gradient descent with STE ensures that the loss is minimized. However, when the gap is large, this guarantee cannot be maintained. We hope to reduce such gradient gap without causing a crash in the codebook.

\begin{figure}[htbp]
    \centering
    \begin{minipage}[t]{0.49\textwidth}
        \centering
        \includegraphics[width=\textwidth]{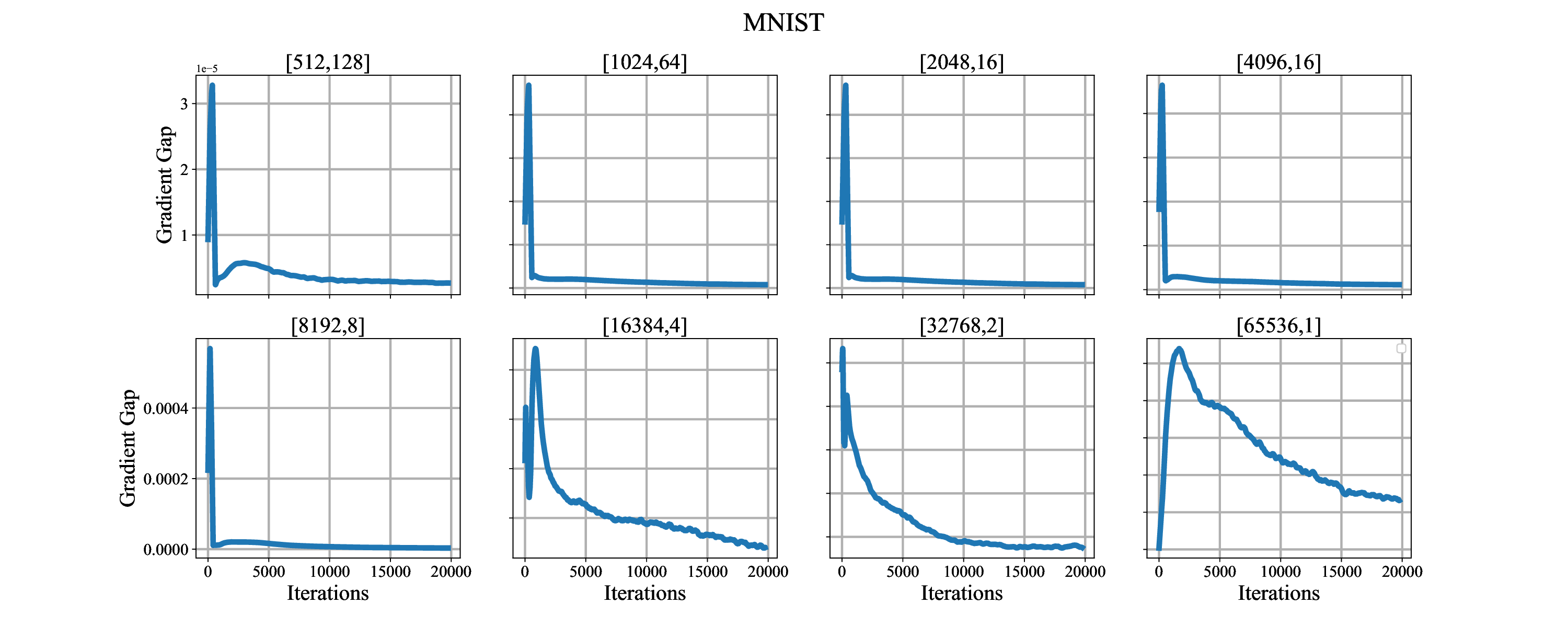}
    \end{minipage}
    \hfill
    \begin{minipage}[t]{0.49\textwidth}
        \centering
        \includegraphics[width=\textwidth]{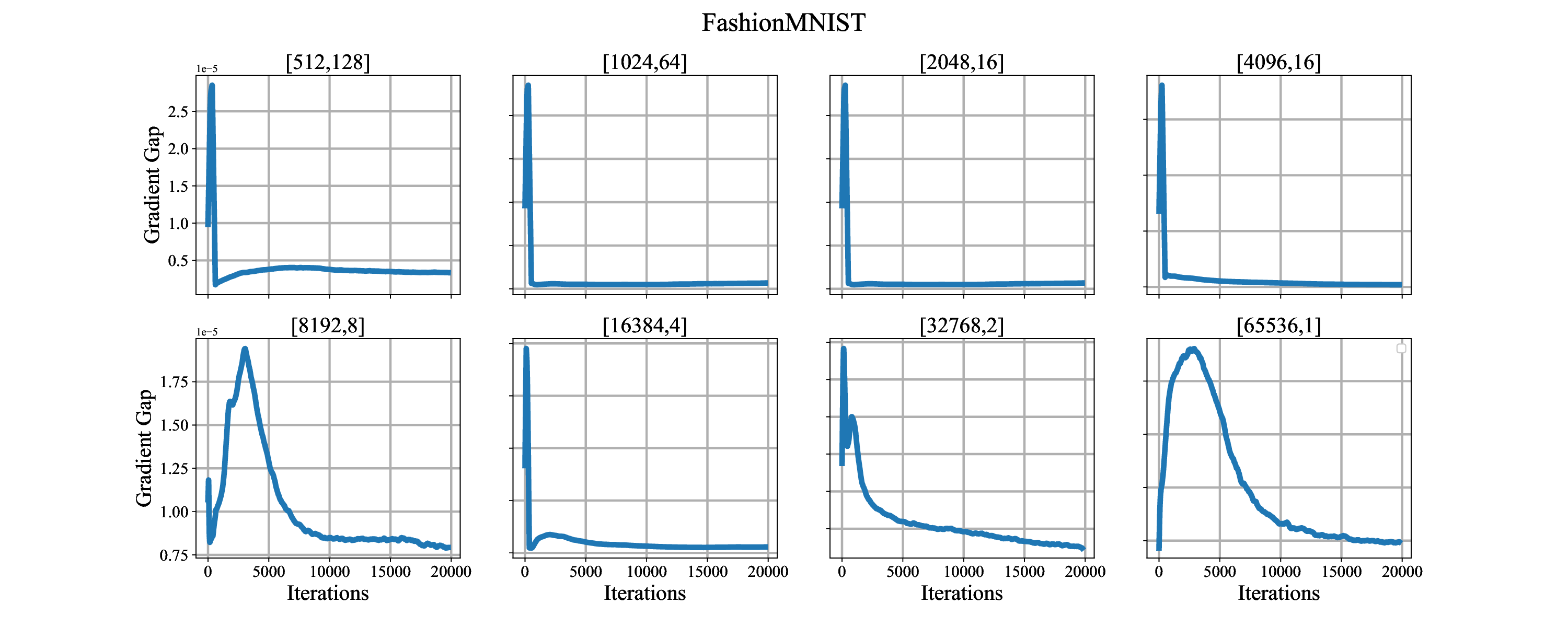}
    \end{minipage}
    
    
    \begin{minipage}[t]{0.49\textwidth}
        \centering
        \includegraphics[width=\textwidth]{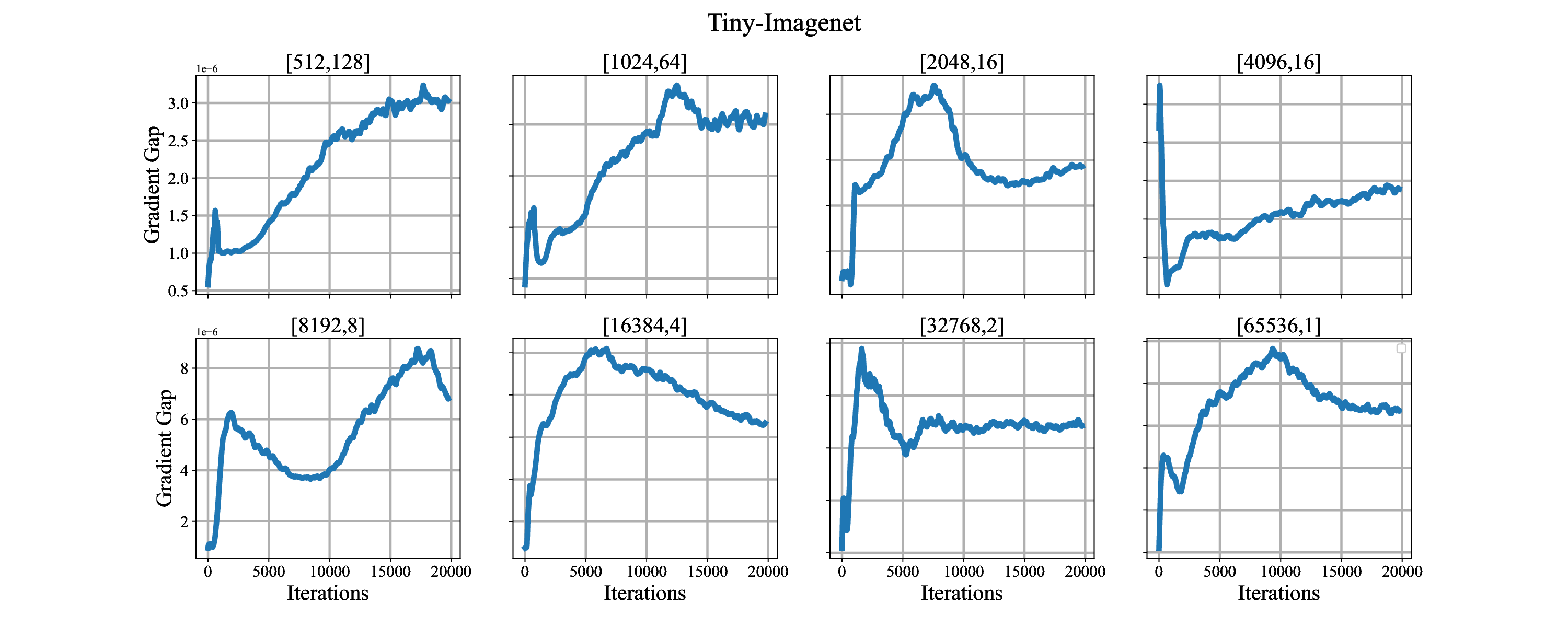}
    \end{minipage}
    \hfill
    \begin{minipage}[t]{0.49\textwidth}
        \centering
        \includegraphics[width=\textwidth]{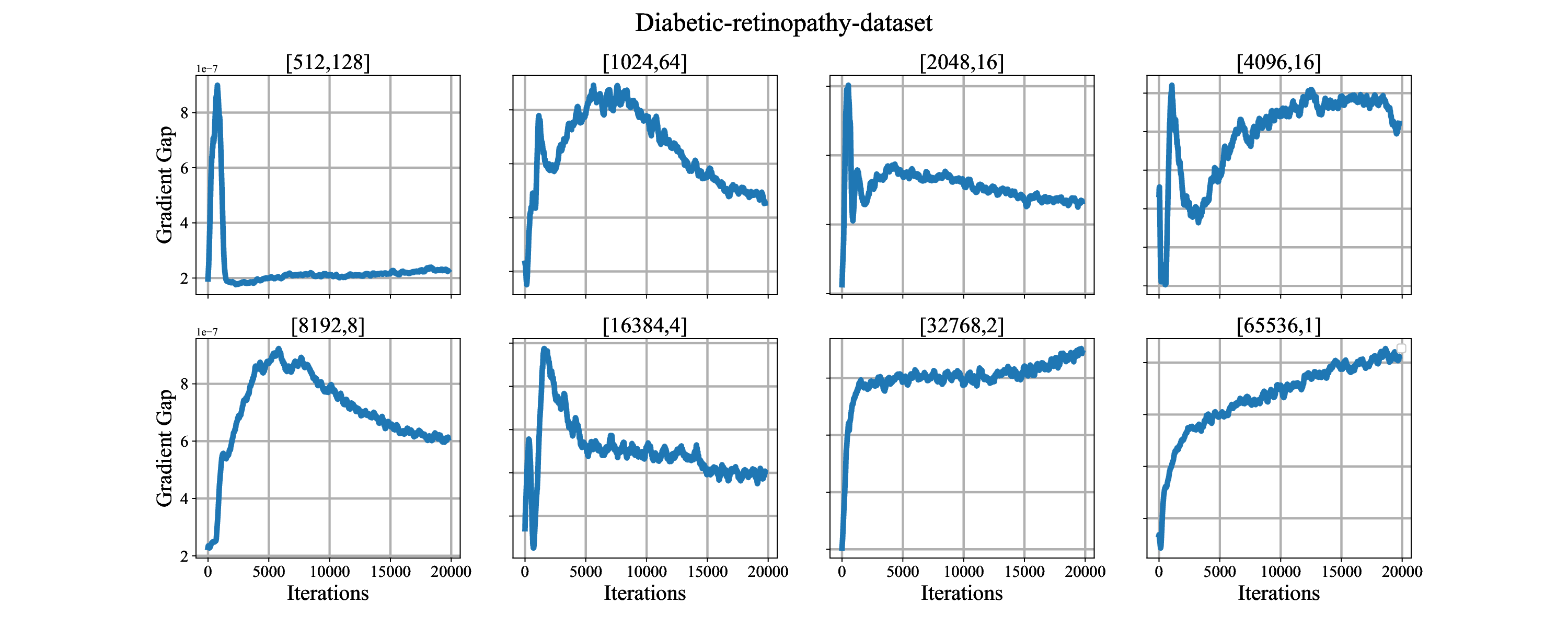}
    \end{minipage}
    \begin{minipage}[t]{0.49\textwidth}
        \centering
        \includegraphics[width=\textwidth]{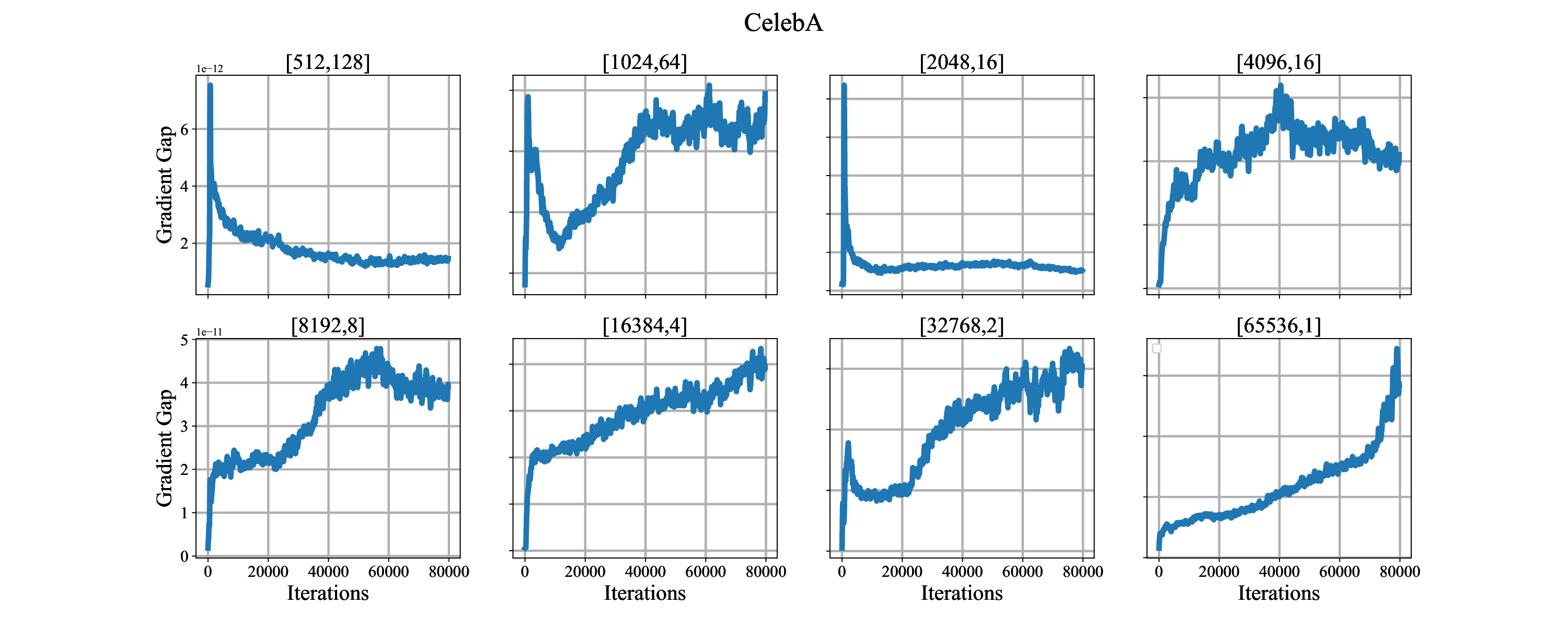}
    \end{minipage}
    \caption{Gadient gap for fixed codebook models. Recording gradient gap variations during training for each dataset under each fixed codebook, with the discrete information space being constant at \(65536\).}
    \label{fig7}
\end{figure}

\begin{figure}[htbp]
    \centering
    \begin{minipage}[t]{0.49\textwidth}
        \centering
        \includegraphics[width=\textwidth]{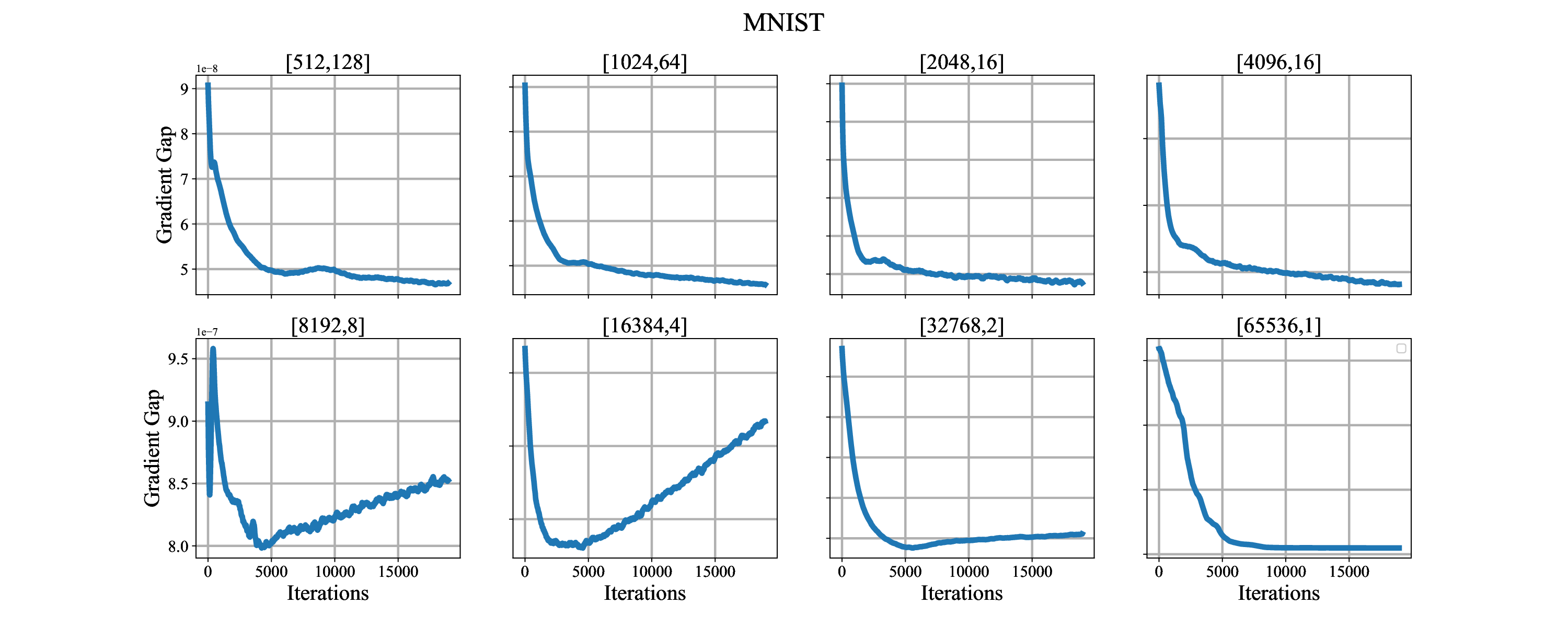}
    \end{minipage}
    \hfill
    \begin{minipage}[t]{0.49\textwidth}
        \centering
        \includegraphics[width=\textwidth]{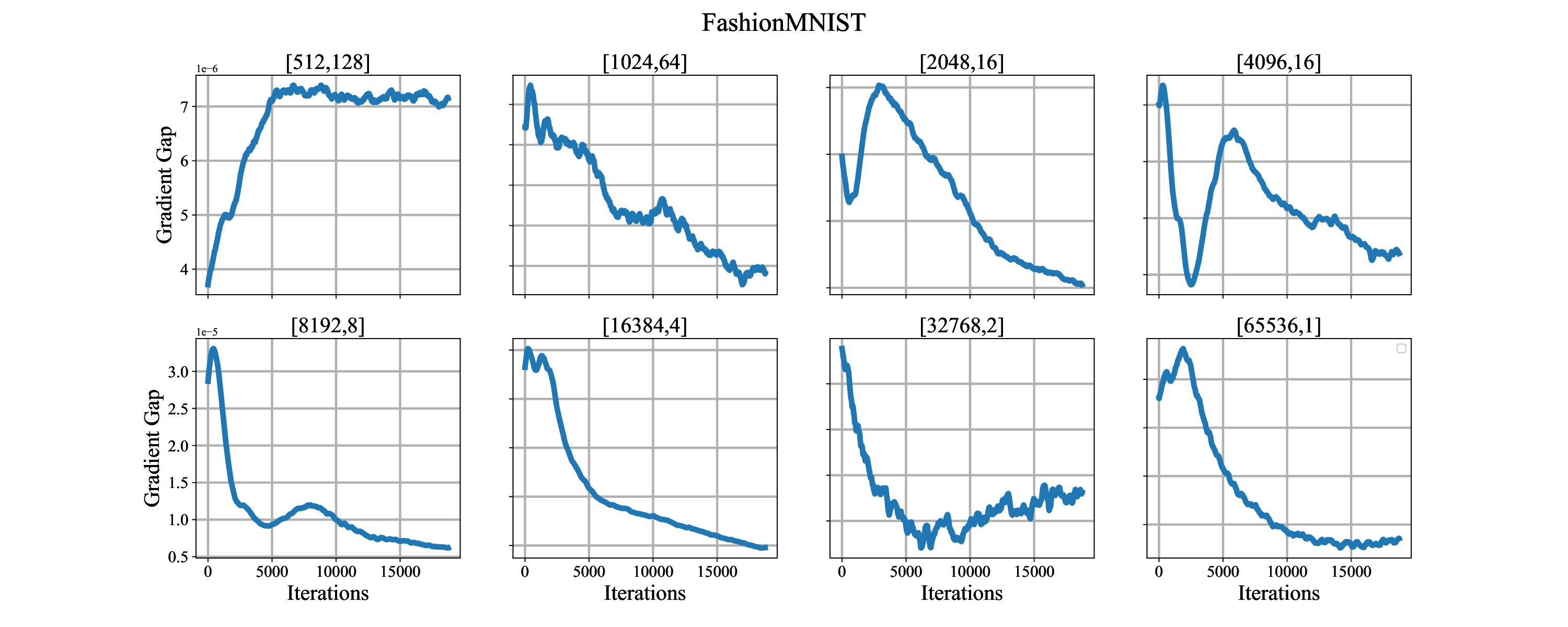}
    \end{minipage}
    \caption{The gradient gap variations of the fixed codebook VQ-VAE model on single-channel datasets.}
    \label{fig8}
\end{figure}

Figure \ref{fig7} illustrates the variation of gradient gap for fixed codebook models across several datasets. Combining the experimental results provided for CIFAR10 in Section \ref{sec4}, we observe that the gradient gap differ between single-channel and three-channel datasets. Naturally, we lean towards acknowledging the three-channel dataset. The model has an excessively high gradient gap during the early stages of training on the single-channel dataset. Therefore, we omitted the gradient gap for the first \(1000\) batches of training and plotted the subsequent curves as shown in Figure \ref{fig8}.

\begin{figure}[htbp]
    \centering
    \begin{minipage}[t]{0.49\textwidth}
        \centering
        \includegraphics[width=\textwidth]{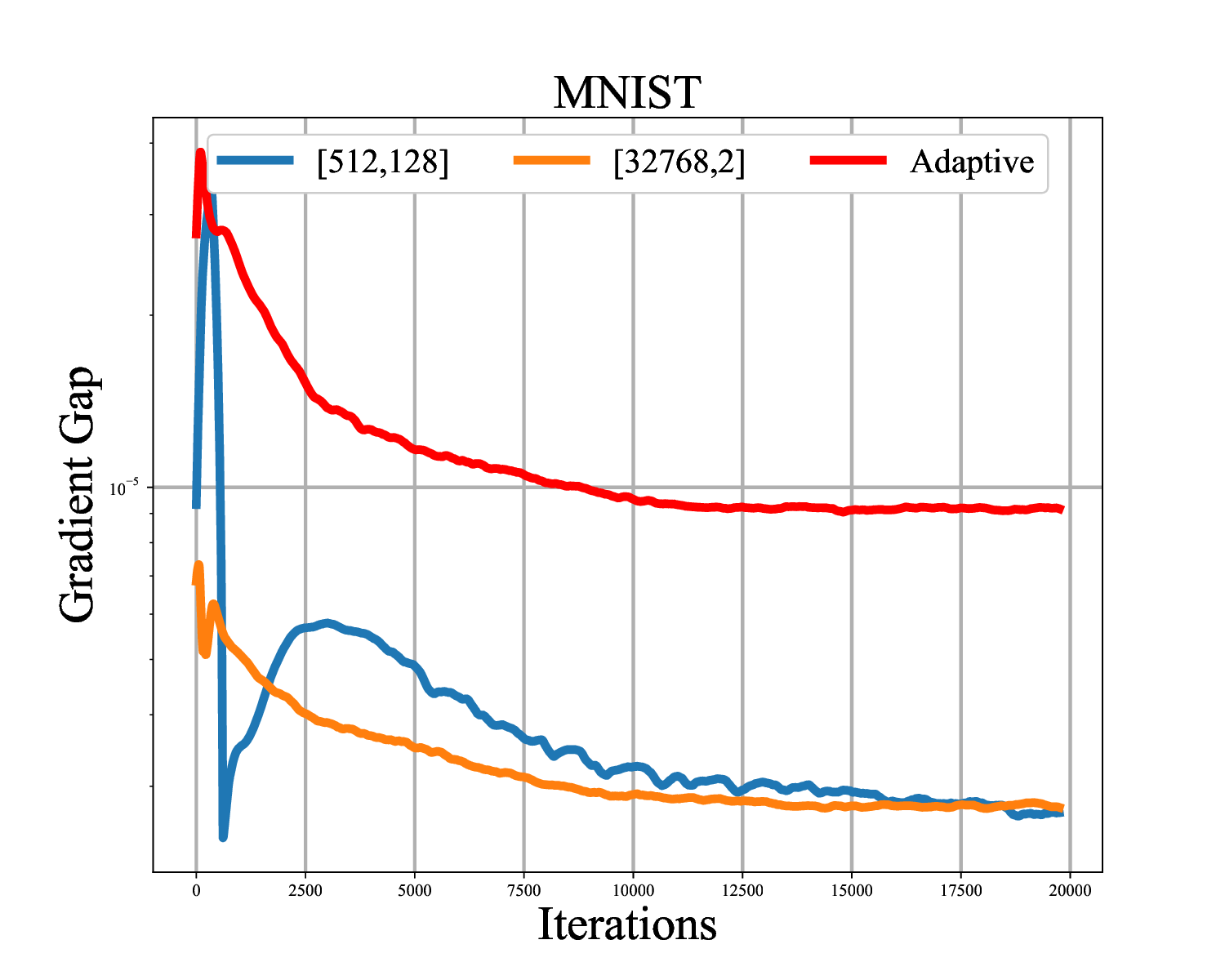}
    \end{minipage}
    \hfill
    \begin{minipage}[t]{0.49\textwidth}
        \centering
        \includegraphics[width=\textwidth]{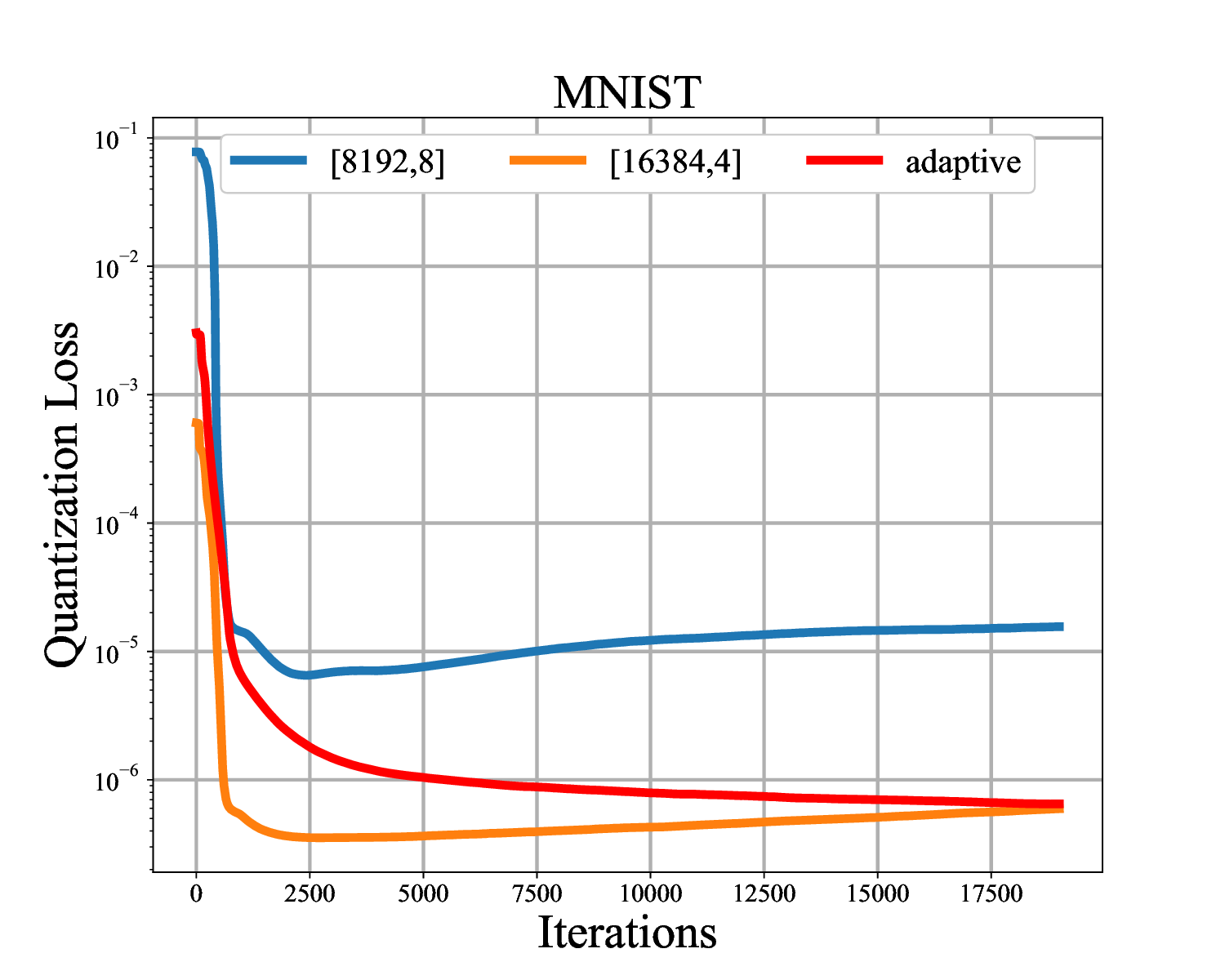}
    \end{minipage}
    \begin{minipage}[t]{0.49\textwidth}
        \centering
        \includegraphics[width=\textwidth]{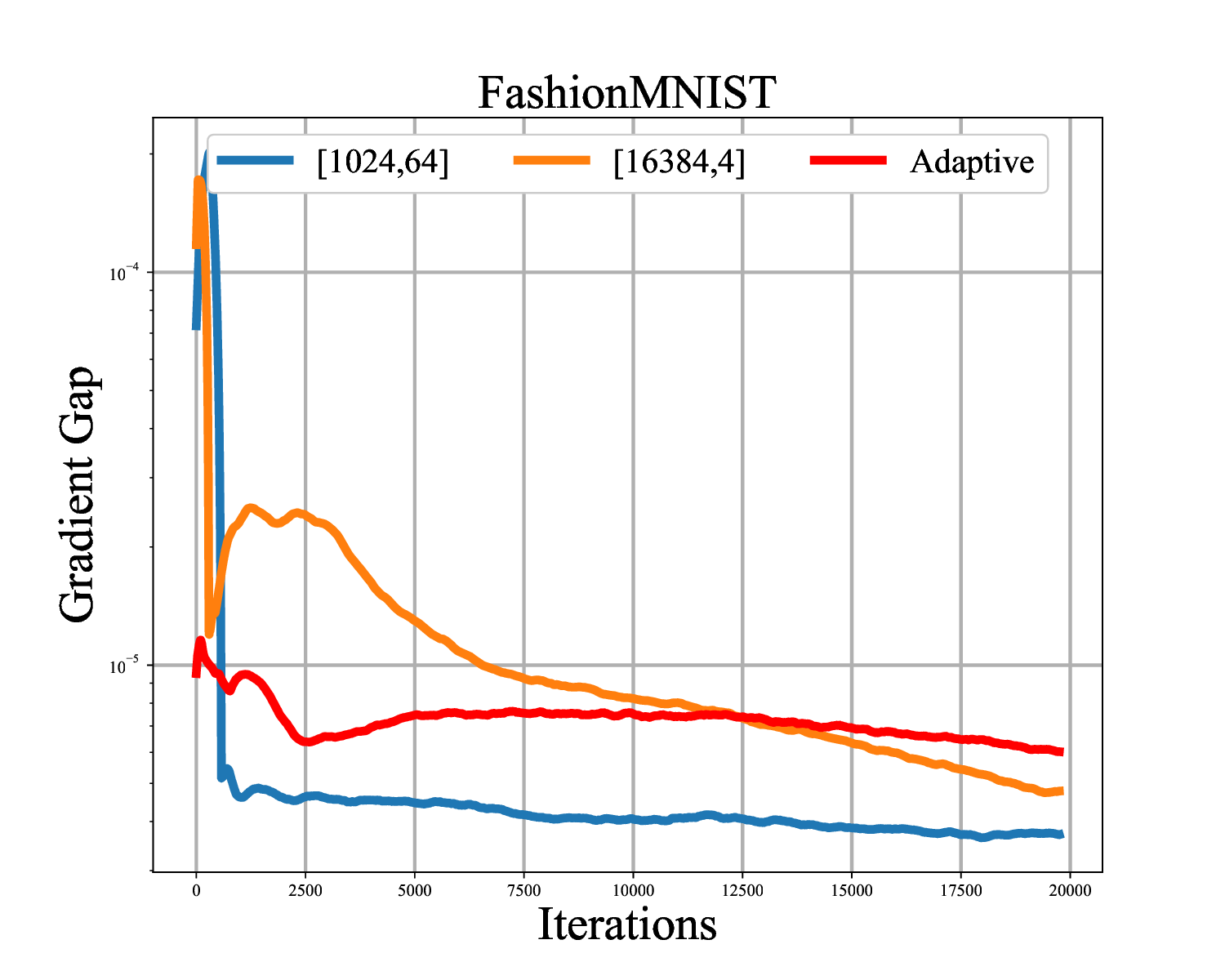}
    \end{minipage}
    \hfill
    \begin{minipage}[t]{0.49\textwidth}
        \centering
        \includegraphics[width=\textwidth]{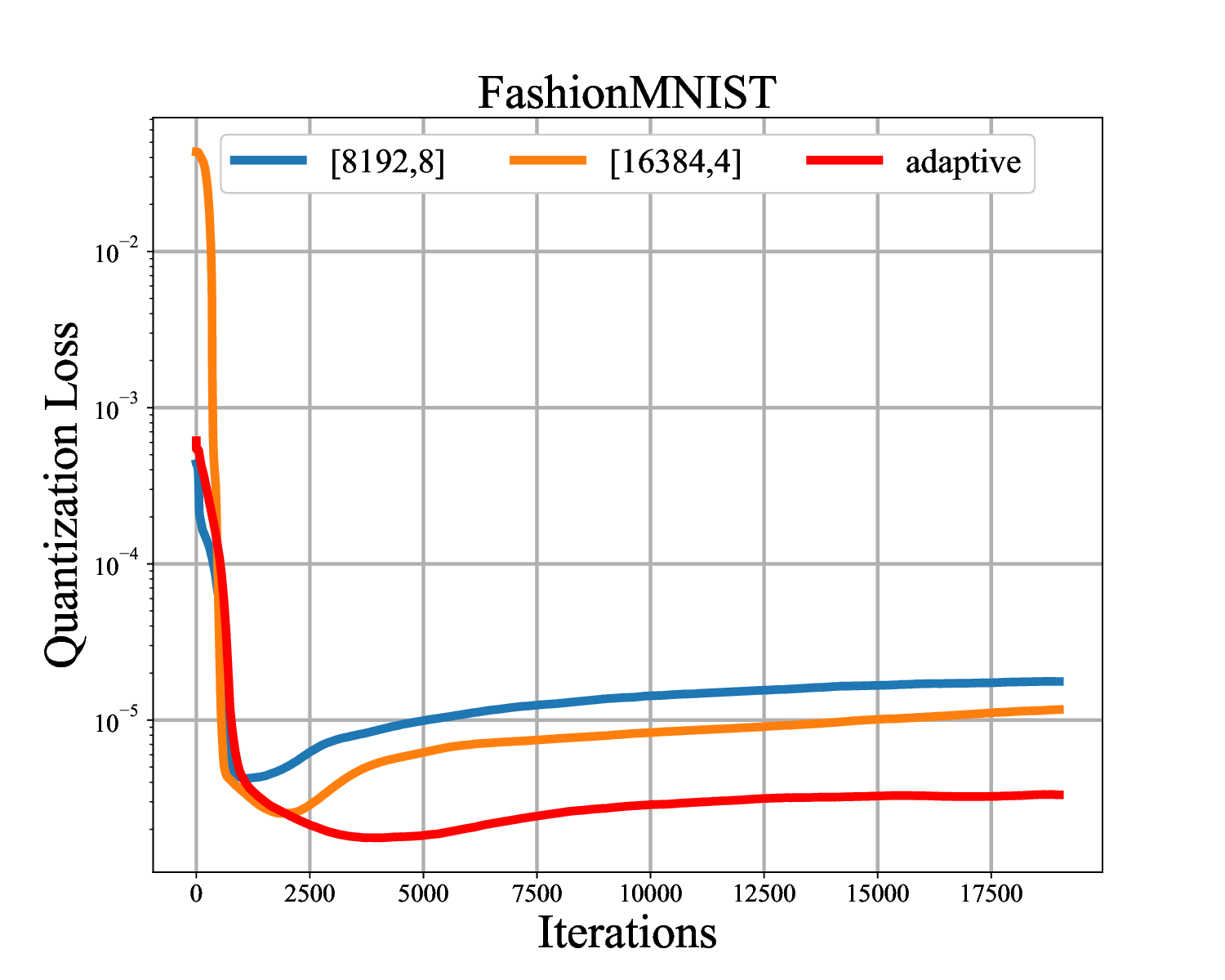}
    \end{minipage}
    \caption{Comparison of gradient gap and quantization loss between the dynamic quantization model and the fixed codebook model. Left: Each plot contains three curves: the blue curve represents the lowest gradient gap in the fixed codebook model, the orange gap represents the gradient gap with the best reconstruction performance in the fixed codebook model, and the red curve represents the gradient gap in the dynamic quantization model. Right: Similar to the left figure, but the data represented is the quantization loss.}
    \label{fig9}
\end{figure}

It can be observed that the gradient gap on single-channel datasets exhibits a decreasing trend throughout the entire training process. However, the gradient gap curve on the three-channel dataset shows less apparent variations. We further illustrate this comparison through Figure \ref{fig9}(Left). On the standard three-channel dataset, the model utilizing dynamic quantization not only exhibits better reconstruction performance compared to the conventional fixed codebook model but also demonstrates low gradient gap. However, such advantages are not as pronounced on single-channel datasets. Due to the relatively small batch size, we speculate that the reason for the different patterns of gradient error variation is the low data complexity of single channel datasets. From Figure \ref{fig9}(Right), it can be seen that the quantization loss of the adaptive dynamic quantization model is similar on all six datasets, approaching the quantization loss of the optimal fixed codebook. Nonetheless, it is evident that our adaptive dynamic approach enhances the reconstruction performance of the VQ-VAE model.
\subsection{Statement}\label{app5}
We guarantee that our research and paper strictly adhere to NeurIPS Code of Ethics.
\subsection{Acknowledgement}\label{app6}
We would like to express our deep gratitude to the authors of the original model VQ-VAE. Their model provides valuable foundation and inspiration for our research. In addition, we would like to thank other relevant researchers for providing code. URL: https://github.com/zalandoresearch/pytorch-vq-vae?tab=MIT-1-ov-file. License: MIT license.

\end{document}